\documentclass[runningheads]{llncs}

\usepackage{wrapfig}
\usepackage{eccv}

\usepackage{eccvabbrv}

\usepackage{graphicx}
\usepackage{booktabs}
\usepackage{multirow}

\usepackage[accsupp]{axessibility}  

\usepackage{hyperref}

\usepackage{orcidlink}

\begin{document}

\title{AFFMAE: Scalable Vision Pre-Training for High-Resolution Microscopy Segmentation on Desktop Hardware} 

\titlerunning{AFFMAE}

\author{David Smerkous\inst{1,3}\thanks{Equal contribution}\orcidlink{0009-0007-3043-1409} \and
Zian (Andy) Wang\inst{1,2}\protect\footnotemark[1]\orcidlink{0009-0005-6526-3690} \and
Behzad Najafian\inst{1}\orcidlink{0000-0002-0904-6721}}

\authorrunning{D. Smerkous and Z. Wang et al.}

\institute{Department of Laboratory Medicine \& Pathology, University of Washington, Seattle WA, USA \and
Paul G. Allen School of Computer Science \& Engineering, University of Washington, Seattle WA, USA \and
Electrical Engineering \& Computer Science, Oregon State University, Corvallis, OR, USA}

\maketitle

\begin{abstract}
Self-supervised pretraining has transformed computer vision by enabling data-efficient fine-tuning, yet high-resolution pretraining typically requires server-scale infrastructure, limiting custom in-domain training for many research laboratories. Masked Autoencoders (MAE) reduce computation by encoding only visible tokens, but combining MAE with hierarchical downsampling architectures has remained structurally challenging due to dense grid priors and mask-aware design compromises. We introduce \textbf{AFFMAE}, a masking-friendly hierarchical pretraining framework built on adaptive, off-grid token merging. AFFMAE removes dense-grid assumptions while preserving hierarchical scalability during pre-training and fine-tuning. To support this architecture, we developed numerically stable mixed-precision Triton kernels and a lightweight, point-based decoder that can be directly repurposed as a segmentation head. On high-resolution microscopy segmentation, AFFMAE matches MAE finetuning performance on foot process width estimation with ViT backbone at equal parameter counts while being 2x faster during pre-training and halving peak memory usage. Furthermore, AFFMAE achieves up to 5x throughput speedups fine-tuning at the $1024$px resolution, providing high-resolution model training on desktop hardware. Code available at \href{https://github.com/najafian-lab/affmae}{https://github.com/najafian-lab/affmae}.
  \keywords{Masked Autoencoders \and Efficient Vision Transformers \and Microscopy Segmentation}
\end{abstract}

\section{Introduction}
\label{sec:intro}

Biomedical diagnostics accumulate large volumes of unlabeled high-resolution images (e.g., histology, electron microscopy, radiology), yet pretraining models remains out of reach for many groups. Two constraints dominate in practice: (i) \textit{compute}, since high-resolution pretraining often assumes server-grade multi-GPU setups, and (ii) \textit{data governance}, where moving sensitive datasets to external cloud infrastructure can trigger lengthy HIPAA/IRB processes and institutional review. These frictions are especially pronounced in \emph{novel} biomedical regimes—e.g., a newly studied cell phenotype, stain, scanner, or acquisition protocol—where no curated public dataset exists, but a lab may still have a large private archive. In such settings, prior work repeatedly finds that pretraining on the target domain improves transfer over generic ImageNet-style initialization \cite{radimagenet,modelgenesis,retfound}; more broadly, transfer improves as the pretraining distribution approaches the downstream domain \cite{domainspecifictransfer}. This motivates \emph{in-house} pretraining methods that are feasible on desktop-class GPUs for privacy-sensitive and institutionally constrained labs.

\begin{figure*}[t]
  \centering
  \includegraphics[width=0.7\textwidth]{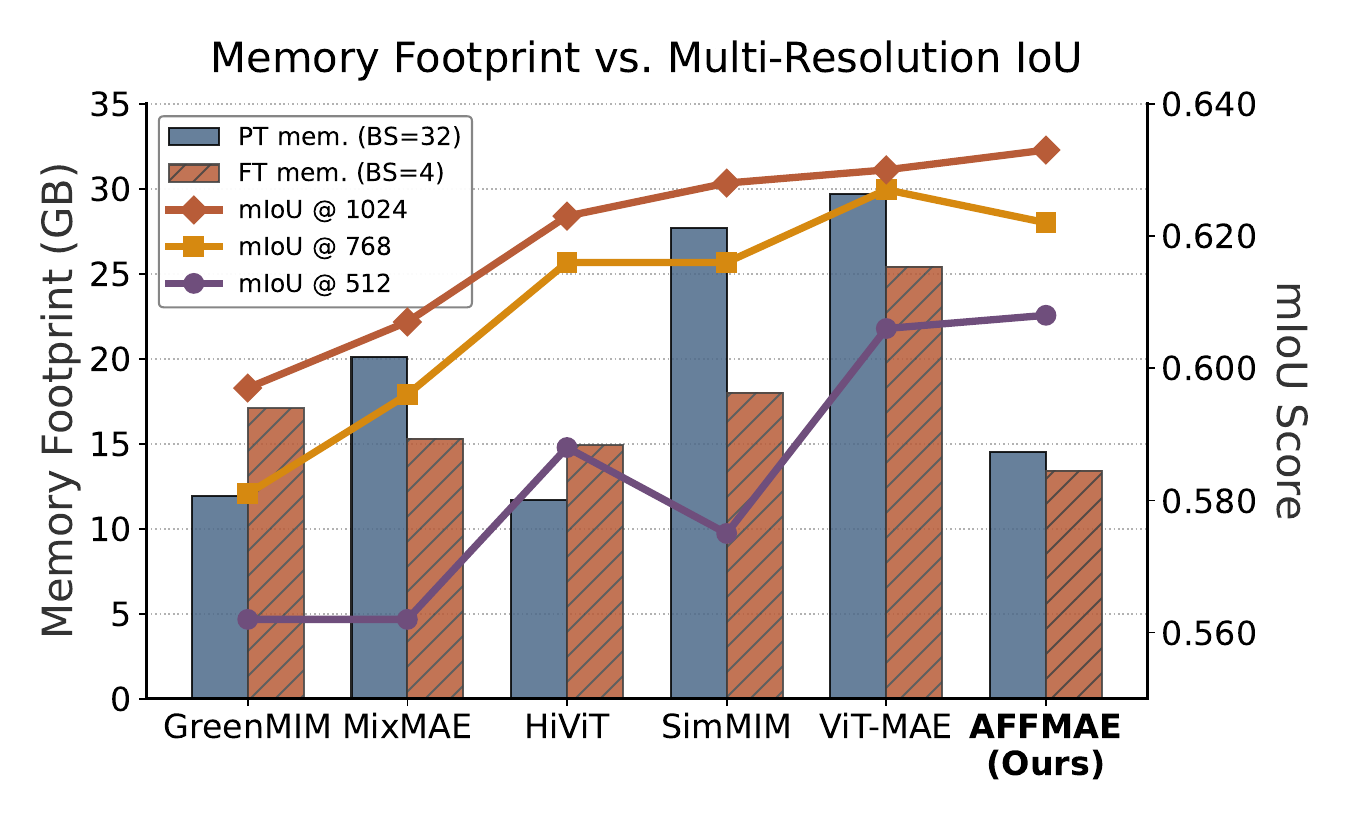}
  \caption{Performance versus computational cost across architectures and resolutions. PT and FT means Pre-Training and Fine-Tuning, respectively. AFFMAE achieves ViT-level segmentation accuracy while requiring significantly smaller memory footprint.}
  \label{fig:main}
\end{figure*}

In this work we focus on kidney EM, where high-resolution, in-domain pretraining is particularly valuable and difficult to scale. Our primary focus are podocytes, which are post-mitotic cells that play a crucial role in maintaining plasma filtration in the kidneys, and most diseases causing end-stage kidney disease are related to podocyte injury. These cells form interdigitated foot processes that widen upon injury, leading to protein leakage into the urine; thus, foot process width (FPW) is a widely used biomarker of podocyte injury in clinical and research settings. EM segmentation for FPW measurement is particularly demanding because it requires both \emph{fine detail} and \emph{global context}: FPW depends on segmenting tiny filtration slits that lie along the surface of the glomerular basement membrane (GBM), and these structures are only reliably resolved at high input resolutions ($>512\times512$) \cite{forknet, laudon2025digital, yamashita2025ai}. Many slit-like textures are locally ambiguous (e.g., within the capillary lumen), so correct classification depends on global glomerular geometry and whether a candidate lies on the GBM boundary; this is further amplified in foot process effacement, where slit morphology is degraded and local neighborhoods become insufficient to identify slits \cite{forknet}. Importantly, these targets form long, thin boundaries, making them especially sensitive to coarse grid-aligned downsampling that can dilute or erase narrow structures (Figure~\ref{fig:qualitative_comparison}), motivating architectures that can retain high spatial detail, while still scaling efficiently to high resolution for global context.

Masked image modeling (MIM), and in particular Masked Autoencoders (MAE), is attractive for resource-constrained in-house pretraining because the encoder runs only on visible tokens, reducing activation memory and FLOPs \cite{he2022masked}. However, ViT-MAE can still be slow and memory-heavy at microscopy-scale resolutions on consumer GPUs, motivating hierarchical backbones that reduce token counts via downsampling. Unfortunately, most hierarchical ViTs (e.g., Swin) are tied to a dense lattice through windowing and patch merging, so discarding tokens breaks their core operations and often forces dense bookkeeping or mask-aware encoders, introducing pretrain--finetune mismatch and degraded downstream performance \cite{xu2023swin,liu2023mixmae,greenmim}.

We address this gap with \textbf{AFFMAE}, a masking-friendly framework for \emph{hierarchical} pretraining that is explicitly designed to make in-domain, in-house pretraining practical for consumer hardware. Building on AutoFocusFormer \cite{ziwen2023autofocusformer}, AFFMAE performs adaptive, off-the-grid dynamic downsampling while keeping the encoder mask-agnostic. The key idea is to perform token merging and attention only over the \emph{visible} tokens, preserving MIM efficiency while retaining the computational benefits of hierarchy without architectural differences between pretraining and finetuning beyond input token masking, and focusing computation on information-rich regions. Additionally, we provide efficient Flash cluster-attention Triton kernels that run on most modern desktop GPUs, not requiring CUDA \cite{cuda}, enabling base-model pretraining on consumer hardware. Empirically, AFFMAE matches ViT-MAE segmentation performance at the same parameter count while using 1/4th of the FLOPs, 1/2 the memory, and training $2\times$ faster on an RTX~5090.

\section{Related Works}

\textbf{Self-supervised pretraining objectives and trade-offs.}
Self-supervised learning for vision spans contrastive objectives (e.g., SimCLR, MoCo) \cite{simclr,moco}, multimodal contrastive objectives (e.g., CLIP) \cite{radfordclip}, and teacher--student / self-distillation approaches (e.g., BYOL, DINO, I-JEPA) \cite{byol,dino,ijepa}. These families produce strong transferable representations but often rely on multi-view pipelines, large effective batch sizes (or queues/memory banks), and/or momentum teachers, increasing memory, compute, and engineering overhead. In contrast, \textbf{masked image modeling (MIM)} methods \cite{bao2021beit,simmim} learn from a within-image reconstruction signal, and \textbf{Masked Autoencoders (MAE)} \cite{he2022masked} discard masked tokens and run the encoder only on visible tokens, making encoder-side cost scale with the visible token count and enabling practical in-domain pretraining on desktop-class hardware.

\textbf{Hierarchical Transformers for efficient high-resolution vision.}
A major challenge for high-resolution imagery is that standard Vision Transformers \cite{vit} incur quadratic attention cost in the number of tokens \cite{vaswani2017attention}. Hierarchical Transformers mitigate this by progressively reducing token counts through downsampling mechanisms such as patch merging and local attention (e.g., Swin \cite{swin}, PVT \cite{pvt}, and Multiscale ViTs \cite{multiscalevisiontransformer}), substantially improving scaling to high-resolution inputs. However, these designs typically assume a \emph{dense, rigid lattice} of tokens to support window partitioning and grid-aligned merging.

\textbf{Masked pretraining for hierarchical Transformers.}
Adapting MIM to hierarchical backbones is structurally non-trivial because many hierarchical operations expect dense token layouts. SimMIM \cite{simmim} works well with hierarchical backbones but typically retains a dense token stream with mask tokens, departing from MAE's efficiency premise of discarding masked tokens in the encoder; SwinMAE \cite{xu2023swin} similarly must reconcile window computation with masking. More recent approaches target MAE-style token dropping in hierarchical settings---MixMAE \cite{liu2023mixmae} mixes visible patches across images, GreenMIM \cite{greenmim} packs sparse visible tokens into dense windows during pre-training, and HiViT \cite{hivit} serializes masked units via lightweight token mixing before a chosen main stage---but they remain coupled to dense-grid assumptions and/or introduce stage- or masking-specific interfaces, contributing to pretrain--finetune discrepancies and weaker performance on small, thin structures compared to ViT-style baselines in our setting.

\section{Methods}
\label{sec:affmae}

\subsection{AutoFocusFormer}
AutoFocusFormer (AFF) \cite{ziwen2023autofocusformer} is a hierarchical backbone that replaces rigid, grid-aligned downsampling (e.g., stride-2 pooling or patch merging) with adaptive downsampling over an irregular set of token locations. Starting from a dense patch embedding, AFF maintains for each token both a feature vector and a 2D coordinate, and performs local computation by building \emph{balanced neighborhoods} over these coordinates: token locations are grouped into equal-size clusters, via a lightweight space-filling-curve procedure, to preserve some linear ordering for attention and faster KNN grouping. AFF neighborhood construction enables local self-attention for a fixed neighborhood size while remaining agnostic to the underlying grid. Within each neighborhood, attention uses relative position derived directly from the 2D token coordinates, allowing computation to follow the token set as it becomes increasingly irregular \cite{ziwen2023autofocusformer}.

After several local-attention blocks, AFF downsamples using a learnable \emph{neighborhood merging} module. An importance score is learned for each token via a one-layer MLP, and a user-specified downsampling rate $d_s\in(0,1]$ controls how many tokens are carried forward to the next stage (e.g., $d_s{=}0.25$ retains 25\% of tokens). The top-ranked tokens are kept as \emph{anchors}, and the remaining tokens are merged into nearby anchors through a differentiable, location-aware aggregation operator that transfers both features and spatial support, yielding a smaller token set with updated coordinates and features. By supporting arbitrary retention ratios (rather than fixed factors like $1/2, 1/4, 1/8$), AFF can preserve high token density in information-rich regions while aggressively merging texture-less areas (Figure~\ref{fig:token_locations}). Crucially, unlike standard grid-based pooling that uniformly downsamples the spatial resolution of the entire feature map, this irregular downsampling reduces the overall token count while strictly retaining the original, high-resolution spatial coordinates around critical structures; we refer the reader to the original AFF paper for full algorithmic details and additional visualizations \cite{ziwen2023autofocusformer}.

\subsection{AFFMAE}
To maximize pre-training efficiency, our framework preserves the asymmetric design of Masked Autoencoders \cite{he2022masked}: we do not introduce mask tokens or modify the encoder downsampling.

To better exploit off-the-grid multi-scale encoder features, we use a point-based deformable cross-attention decoder \cite{zhu2020deformable}. Decoder queries are learnable mask tokens with absolute positional encodings. For each query with feature $\mathbf{f}$ and reference location $\mathbf{r}\in\mathbb{R}^2$, we predict an offset
\[
\Delta=g(\mathbf{f}),\qquad \mathbf{q}=\mathbf{r}+\Delta,
\]
where $g(\cdot)$ is a linear projection. At each encoder stage $\ell$, we gather the $K$ nearest visible tokens to $\mathbf{q}$ (KNN in 2D coordinate space) to form a \emph{virtual sampled token}. Unlike bilinear sampling used in point-based decoders such as Mask2Former \cite{cheng2022masked}, AFF \cite{ziwen2023autofocusformer} aggregates neighbors with inverse-distance weighting:
\[
d_i=\|\mathbf{q}-\mathbf{x}_i\|_2+\varepsilon,\qquad \tilde{w}_i=d_i^{-p},\qquad w_i=\frac{\tilde{w}_i}{\sum_{j=1}^{K}\tilde{w}_j},
\]
where $p$ controls the spatial concentration of the interpolation. Directly computing the inverse-power weights can be numerically fragile under mixed precision: for small distances, $d_i^{-p}$ can become very large, whereas for more distant neighbors it can underflow toward zero, producing saturated or unstable normalizations. To improve numerical stability, we replace the inverse-power kernel with an exponential distance kernel,
\[
\tilde{w}_i=\exp(-p\,d_i),\qquad w_i=\frac{\exp(z_i - z_\text{max})}{\sum_{j=1}^{K}\exp(z_j - z_\text{max})}=\mathrm{softmax}_i(-p\,d_i),
\]
where $z_i=-p d_i$ and $z_{\max}=\max_k z_k$. This yields the standard softmax form, allowing the weights to be evaluated with max-logit subtraction for stable mixed-precision computation while retaining an explicit locality parameter $p$. The resulting virtual sampled token is
\[
\hat{\mathbf{f}}^{(\ell)}(\mathbf{q})=\sum_{i=1}^{K}\mathrm{softmax}_i\!\left(-p\,d_i\right)\,\mathbf{f}_i^{(\ell)}.
\]

Our decoder comprises four stages, directly mirroring the four hierarchical stages of the encoder. At each stage, deformable cross-attention is applied against the corresponding stage-wise virtual tokens, followed by deformable self-attention. Following the final cross-attention stage, the dense token representations are processed through a standard LayerNorm and a linear reconstruction head to predict the original pixel values.

Because this decoder natively restores hierarchical spatial structures back on a dense grid, it is sufficient to be directly repurposed for downstream segmentation tasks with competitive performance as FPN-based segmentation decoders as shown in Section \ref{sec:experiments}. This avoids the complexity of adapting and tuning separate, heavy task-specific heads (e.g., UperNet or Mask2Former) during fine-tuning.

\subsection{Flash-style Cluster Attention and Fast Decoder KNN}
While AFF reduces FLOPs by operating on sparse neighborhoods, the original implementation does not translate into proportional wall-clock speedups: cluster attention explicitly materializes attention scores (computing $QK^\top$ and storing the score matrix before multiplying by $V$), which is memory-bound and can be numerically unstable under mixed precision. We therefore re-implement local cluster attention in Triton following the FlashAttention principle \cite{flashattn}: we \emph{stream} attention computation over tiles, fuse the neighborhood attention steps, and accumulate the softmax normalization in fp32 while writing outputs in fp16. Concretely, for each neighborhood $n$ we compute
\[
\mathrm{FlashNbhdAttn}(Q,K,V)=\mathrm{softmax}\!\big(Q_n [K_n \;\; K_{\text{blank}}]^\top + B_n\big)\,[V_n \;\; V_{\text{blank}}],
\]
without materializing the full score matrix, where $(K_\text{blank},V_\text{blank})$ are learned blank tokens used to regularize attention norms in texture-less regions as in \cite{ziwen2023autofocusformer}. This reduces memory traffic, improves throughput, and improves half-precision stability via a numerically stable streaming softmax with fp32 accumulators.

In the decoder, a major overhead comes from repeatedly retrieving the $K$ nearest visible tokens for each continuous query location $\mathbf{q}$ when constructing virtual sampled tokens. Instead of performing a separate KNN search for every query, requiring a full pass over all visible token locations, we exploit the fact that token coordinates originate from a dense $H \times W$ patch grid (e.g., $64 \times 64 = 4096$ locations for a $512^2$ image with patch size $8$). We precompute a compact $H\times W \times K$ lookup table over this grid, where each entry stores the indices and 2D coordinates of the nearest visible-token candidates. At runtime, each continuous query location is quantized to its nearest grid cell, and neighbor retrieval becomes a table gather followed by a small fixed-size distance computation. After table construction, this makes KNN retrieval effectively $O(1)$ per query and allows the distance computation and softmax weighting to be fused into the same Triton kernel used for virtual token construction. This cache-based lookup reduces decoder overhead and results in a $1.5\times$ forward-time speedup and a $1.4\times$ backward-time speedup in the point decoder (Supp.~1.1).

\subsection{Deep Supervision}

\begin{figure}[ht]
    \centering
    \begin{subfigure}[c]{0.48\linewidth}
        \centering
        \includegraphics[width=\linewidth]{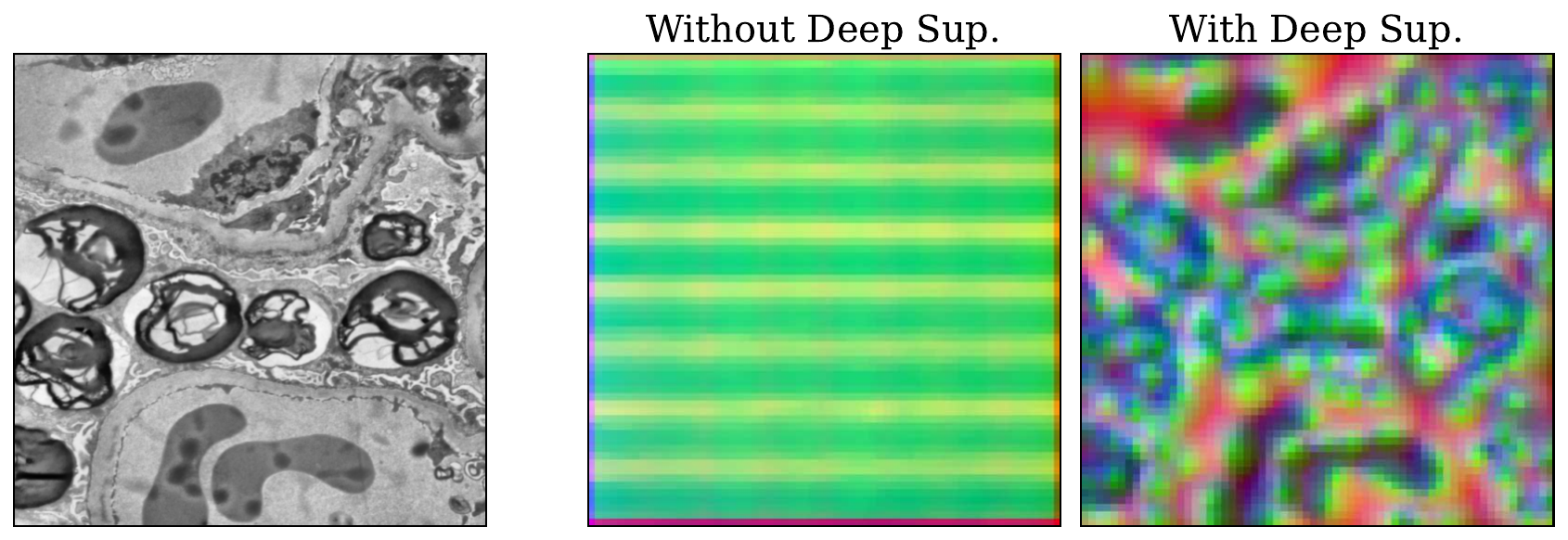}
        \caption{\textbf{PCA projections of the learned representations.} Features were extracted immediately after cross-attention with the respective encoder stages, allowing us to visualize a dense spatial grid instead of scattered visible tokens. The green PCA circles correspond with encoder downsampled token locations.}
        \label{fig:deep_sup_pca}
    \end{subfigure}
    \begin{subfigure}[c]{0.48\linewidth}
        \centering
        \includegraphics[width=\linewidth]{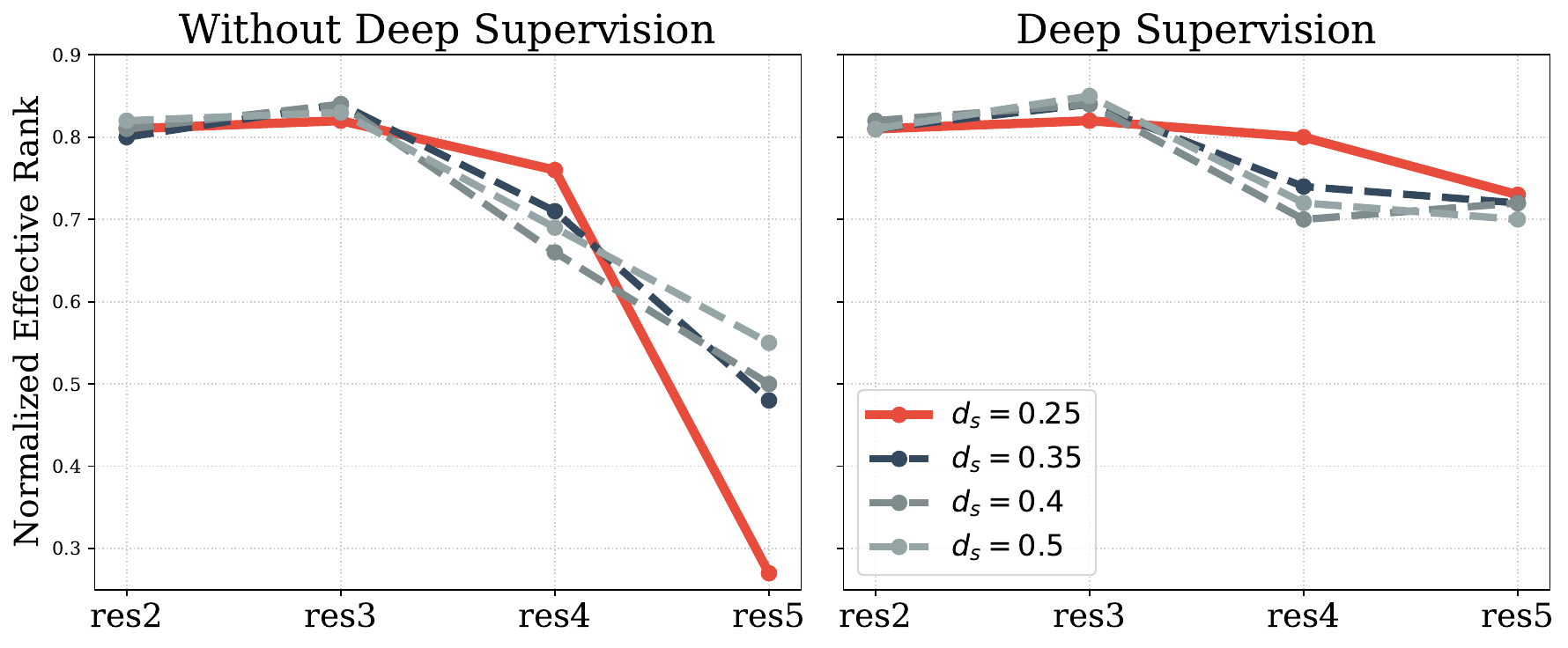}
        \caption{\textbf{Normalized Effective Rank.} The evolution of representation rank across layers, computed from the encoder tokens.}
        \label{fig:rank_collapse}
    \end{subfigure}
    \hfill
    \caption{Analysis of representational collapse and the effect of Deep Supervision.}
    \label{fig:deep_supervision_combined}
\end{figure}

When adapting AFF to MIM-style pre-training with aggressive downsampling, we empirically observe a severe degradation in the representational quality of the deepest encoder stages. Principal Component Analysis (PCA) visualizations of the latent tokens at the final, sparsest stages reveal a loss of semantic structure. Without intervention, the features degenerate into uniform, grid-like patterns, indicating that the tokens are primarily representing their positional embeddings rather than the underlying image semantics (Figure \ref{fig:deep_sup_pca}).

To mitigate this, we employ Deep Supervision \cite{lee2015deeply}. We attach auxiliary reconstruction heads at the end of each intermediate decoder stage. By forcing the network to reconstruct the masked input directly from these intermediate tokens before they have cross-attended to the dense tokens at the shallowest encoder stage, we inject a strong, localized learning signal that encourages the retention of semantic information in the sparser stages.

We validate the use of Deep Supervision by monitoring the Normalized Effective Rank \cite{effrank} $\hat{R}$ of the feature space. For a batch of tokens at a specific downsampling stage with singular values $\sigma_i$, $\hat{R}$ is the exponential of the spectral entropy normalized by $\min(N,D)$:
\begin{equation}
    \hat{R} = \frac{1}{\min(N, D)} \exp\left(-\sum p_i \log p_i\right), \quad p_i = \frac{\sigma_i}{\sum \sigma_j}
\end{equation}
where $N$ and $D$ represent the number of tokens at that stage and the embedding dimension, respectively. Figure \ref{fig:rank_collapse} visualizes the evolution of $\hat{R}$ across encoder stages. Without Deep Supervision, as the downsampling rate becomes more aggressive, the effective rank plummets at deeper stages; a phenomenon that remains severe even at moderate downsampling rates. Conversely, with Deep Supervision, the network successfully maintains a high effective rank ($\hat{R} > 0.7$) across all stages, largely independent of the downsampling rate.

\subsection{Perlin Masking}
\begin{figure}[tb]
  \centering
  \begin{subfigure}[b]{0.6\textwidth}
    \centering
    \includegraphics[width=\linewidth]{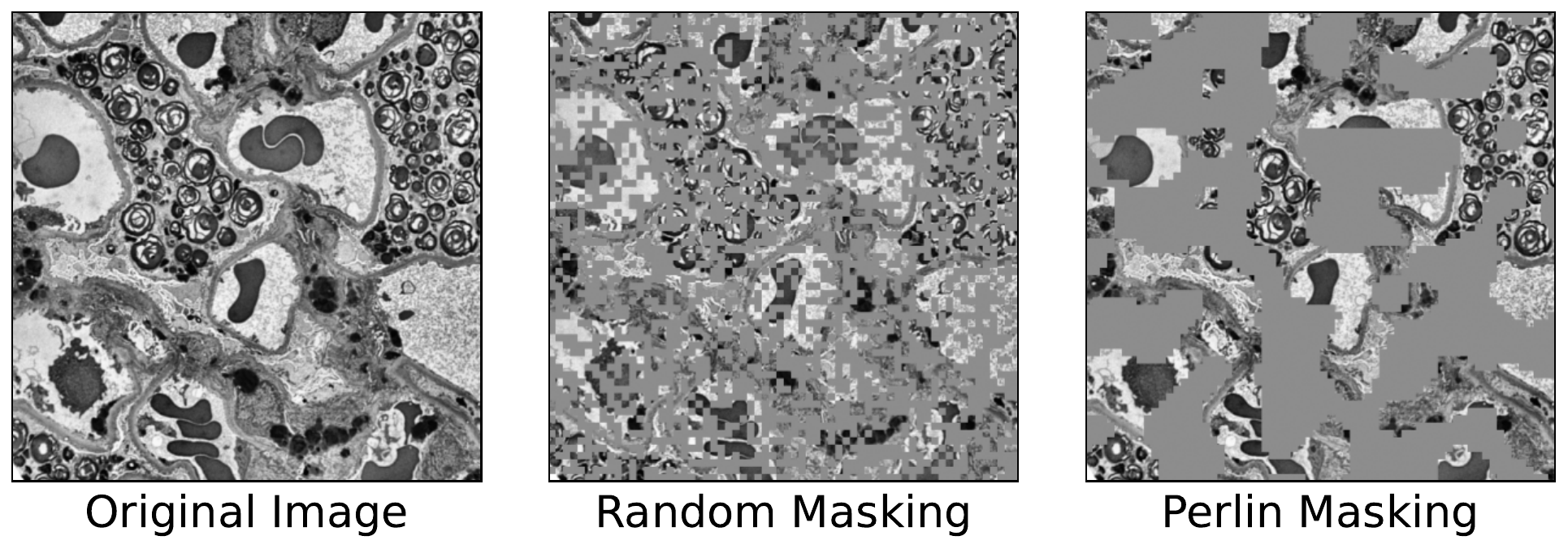}
    \caption{Illustration of our Perlin Masking strategy with a masking ratio of $0.5$ and a patch size of $8$.}
    \label{fig:perlin_mask}
  \end{subfigure}%
  \hfill
  \begin{subfigure}[b]{0.35\textwidth}
    \centering
    \includegraphics[width=\linewidth]{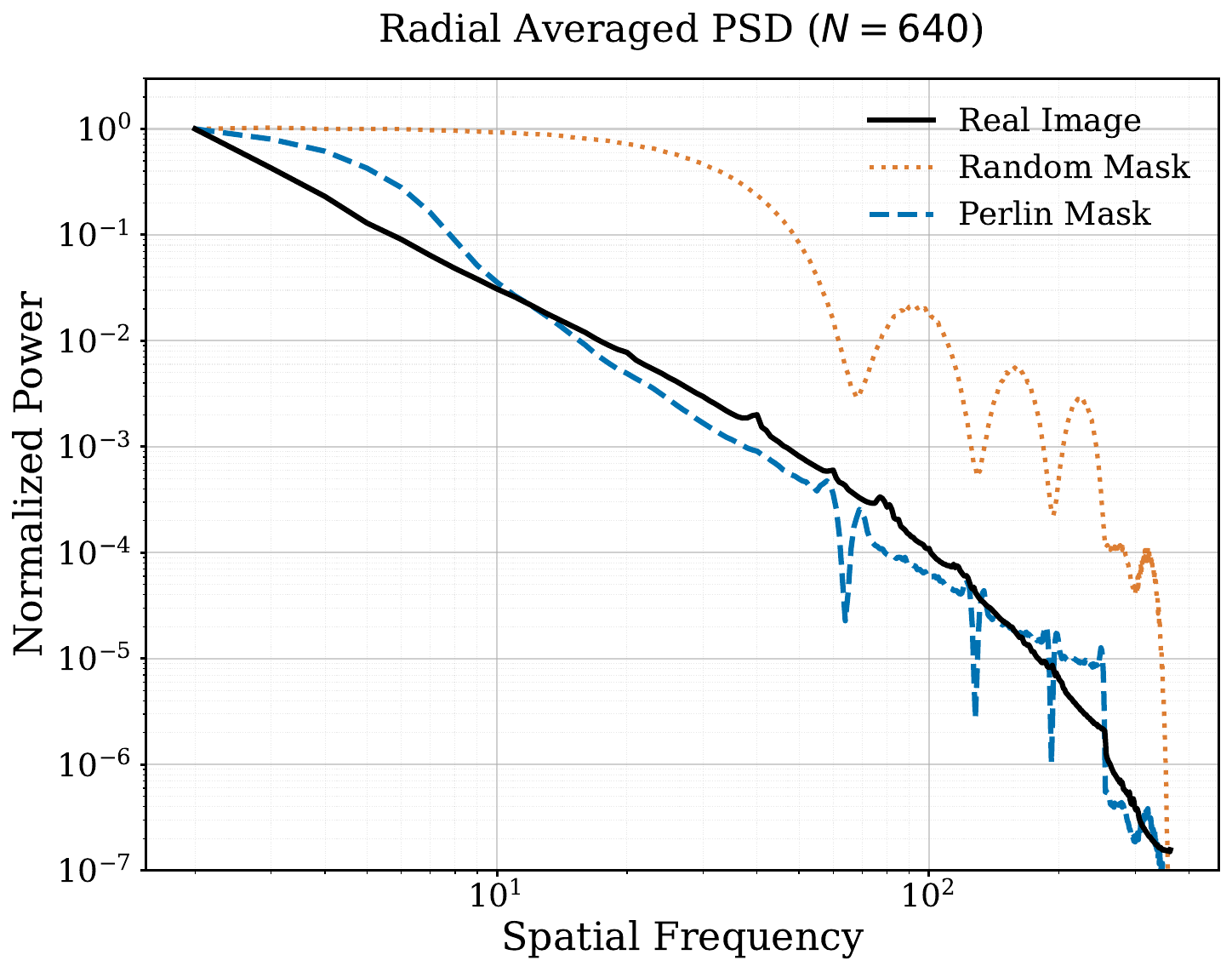}
    \caption{Spectral analysis of masking strategies.}
    \label{fig:psd_analysis}
  \end{subfigure}
  
  \vspace{-0.5em}
  \caption{Visualization and spectral evaluation of the proposed masking approach.}
  \label{fig:combined_masking}
\end{figure}

At high resolutions, random masking becomes increasingly ineffective. As the relative patch size decreases, the model can interpolate missing information from immediate neighbors without capturing the global structure of biological features. We therefore seek a masking strategy that removes contiguous regions analogous to biological structures, so minimizing reconstruction loss requires learning global context and structural coherence. To achieve this, we utilize Perlin noise \cite{perlin1985} to generate masks (Figure~\ref{fig:perlin_mask}).

To justify our masking strategy, we analyze the spectral statistics of the generated masks compared to the statistics of the EM dataset. Natural images, including biological microscopy, exhibit a power spectral density (PSD) that follows a power law decay $S(f) \propto 1/f^\alpha$ (typically $\alpha \approx 2$) \cite{field87, ruderman1994}. We compute the radially averaged PSD for the input images and the binary masks generated by both random and Perlin strategies.

Figure \ref{fig:psd_analysis} presents the results of the spectral analysis. As expected, the original EM images exhibit the characteristic $1/f$ decay typical of natural signals. Standard random masking disrupts these statistics, introducing a high-frequency plateau indicative of white noise artifacts, alongside a significant spectral gap at mid-to-low frequencies. In contrast, the spectral profile of Perlin masking closely tracks both the slope and magnitude of the real biological signal, with barely any gaps over the entire frequency range.

\section{Experiments}
\label{sec:experiments}
\subsection{Experimental Settings}

\textbf{Datasets.} We pre-train on 187,270 unlabeled glomerulus EM images taken between (7500-15000x) and fine-tune on the Foot Process Width (FPW) dataset (570 train, 235 test images) to segment the Podocyte Glomerular Basement Membrane Interface (PGBMI), Foot Process Filtration Slits (Slits), and background. To evaluate the transferability of AFFMAE and ensure our architectural improvements generalize, we additionally fine-tune on public datasets such as Lucchi++ FIB-SEM mitochondria dataset \cite{casser2020fast}. These evaluations on public microscopy benchmarks are provided in the supplementary material.

\textbf{Implementation Details.} We evaluate our approach against other efficient masked modeling baselines \cite{liu2023mixmae, greenmim, hivit, simmim}, the standard MAE \cite{he2022masked} with a ViT \cite{vit} backbone, and nnUNetV2 \cite{isensee2024nnu}. During pre-training, all backbones are scaled to a $\sim$65M parameter budget. Models are pre-trained for 400 epochs on a single NVIDIA RTX 5090 at $512 \times 512$ resolution and an effective batch size of 256. We fine-tune end-to-end for 400 epochs at 512, 768, and 1024 resolutions using a class-weighted combined BCE and Dice loss on the FPW dataset. All MIM models adopt the UperNet decoder \cite{xiao2018unified, he2022masked} matched to the AFFMAE decoder's parameter count ($\sim$10M) for downstream segmentation. FPW performance is evaluated using mean Intersection over Union (mIoU). Full hyperparameter configurations are detailed in the supplementary material.

\subsection{Ablation Studies}
We ablate key components of our framework—downsampling rate, masking ratio, masking strategy, and deep supervision—using a lightweight 23M parameter AFFMAE variant pre-trained for 300 epochs.

\textbf{Downsampling Rate.} In Table \ref{tab:ablation_ds}, we analyze the impact of the adaptive downsampling rate. As expected, higher token retention ($0.5$) yields the highest mIoU ($0.6009$). However, the performance drop-off at $0.4$ is minimal while the required computation decreases by 16\%, representing a sweet spot for efficiency. Aggressive downsampling ($0.25$) leads to a noticeable performance degradation ($0.5729$), likely due to the lack of tokens to represent spatial details required for segmenting small structures like filtration slits.

\textbf{Masking Ratio.} Table \ref{tab:ablation_mask} investigates the pre-training masking ratio. Performance peaks at $50\%$, but drops off noticeably at $75\%$ ($0.5739$). This contrasts with the standard recommendation for MAE of $75\%$. We hypothesize this is due to the fine-grained nature of biological structures (e.g., thin basement membranes), where aggressive masking ($>50\%$) makes it significantly more difficult for the model to learn detailed, high-frequency features during pre-training.

\textbf{Deep Supervision.} We verify the necessity of Deep Supervision in sparse architectures in Table \ref{tab:ablation_deepsup}. At $d_s=0.4$, removing Deep Supervision causes a sharp performance drop to $0.5734$. Re-introducing it recovers performance to $0.5908$ ($+3\%$). This aligns with our observation in Section \ref{sec:affmae}, confirming that auxiliary losses are critical for preventing feature collapse in deep, sparse layers. Another reasonable approach to resolving feature collapse is to enforce reconstruction solely on the deepest encoder tokens. However, as shown (``Stage 4 Recon.''), this yields poor downstream performance. We believe this is due to the lack of the necessary spatial granularity for tokens at the deepest stage to guide dense semantic learning on their own, especially when fine-tuned for segmentation tasks.

\begin{table}[t]
  \caption{\textbf{Ablation Studies.} We ablate key components of the AFFMAE framework. Default settings (unless otherwise specified) are $d_s=0.4$, $0.5$ masking ratio, Perlin masking, and deep supervision enabled.}
  \label{tab:ablations}
  \centering
  
  \begin{subtable}[t]{0.26\linewidth}
    \centering
    \caption{Downsampling Rate.}
    \label{tab:ablation_ds}
    \setlength{\tabcolsep}{2pt} 
    \begin{tabular}{@{}lcc@{}}
      \toprule
      $d_s$ & GFLOPs & mIoU \\
      \midrule
      0.50 & 36 & \textbf{0.6009} \\
      0.40 & 30 & 0.5908 \\
      0.35 & 28 & 0.5753 \\
      0.25 & 23 & 0.5729 \\
      \bottomrule
    \end{tabular}
  \end{subtable}%
  \hfill
  \begin{subtable}[t]{0.21\linewidth}
    \centering
    \caption{Mask Ratio.}
    \label{tab:ablation_mask}
    \setlength{\tabcolsep}{3pt}
    \begin{tabular}{@{}lc@{}}
      \toprule
      Ratio & mIoU \\
      \midrule
      0.35 & 0.5853 \\
      0.50 & \textbf{0.5908} \\
      0.75 & 0.5739 \\
      \bottomrule
    \end{tabular}
  \end{subtable}%
  \hfill
  \begin{subtable}[t]{0.26\linewidth}
    \centering
    \caption{Deep Sup.}
    \label{tab:ablation_deepsup}
    \setlength{\tabcolsep}{3pt}
    \begin{tabular}{@{}lc@{}}
      \toprule
      Method & mIoU \\
      \midrule
      None & 0.5734 \\
      Stage 4 Recon. & 0.4353 \\
      \textbf{Deep Sup.} & \textbf{0.5908} \\
      \bottomrule
    \end{tabular}
  \end{subtable}%
  \hfill
  \begin{subtable}[t]{0.22\linewidth}
    \centering
    \caption{Masking Strategy. $d_s=0.5$}
    \label{tab:ablation_strategy}
    \setlength{\tabcolsep}{3pt}
    \begin{tabular}{@{}lc@{}}
      \toprule
      Strategy & mIoU \\
      \midrule
      Random & 0.5947 \\
      \textbf{Perlin} & \textbf{0.6009} \\
      \bottomrule
    \end{tabular}
  \end{subtable}
  
\end{table}

\textbf{Masking Strategy.} Table \ref{tab:ablation_strategy} compares standard Random masking against our Perlin masking. Random masking yields $0.5947$, while Perlin masking improves this to $0.6009$, aligning with our expectations that Perlin Masking allows the model to learn more useful structural information than a Random masking strategy.

\subsection{Main Results}

Having established our architectural configuration, we evaluate AFFMAE against the dense MAE and other efficient MIM frameworks. We first highlight the importance of in-domain pre-training by comparing the downstream mIoU (at $512 \times 512$) between the MAE baseline trained from scratch, initialized with official MAE ImageNet-1K weights (ViT-Base, 86M parameters, 1600 epochs), and pre-trained on our in-domain EM dataset:

\begin{center}
  \small
  \begin{tabular*}{0.6\linewidth}{@{\extracolsep{\fill}}ccc@{}}
    Scratch & ImageNet-1K & \textbf{In-Domain (EM)} \\
    \midrule
    0.476 & 0.494 & \textbf{0.606} \\
  \end{tabular*}
\end{center}

Even though standard ImageNet weights provide a boost over random initialization, we note that in-domain pre-training contributes a massive performance improvement. We found that even when comparing against the larger model (86M vs 66M), much larger dataset (1.28M vs 187k), and 4x the epochs (1600ep vs 400ep) the MAE ImageNet-1k pretrained ViT model underperformed the smaller in-domain pre-trained ViT model. Showing the need for efficient pretraining methods for niche domains. All subsequent models in our evaluations are pre-trained on the EM dataset.

\begin{table*}[tb]
  \caption{Main downstream fine-tuning results on the FPW dataset. Results are reported as mean $\pm$ std across $4$ random seeds. Throughput (Img/s) and memory are measured with batch size $4$ to ensure OOM-free evaluation across resolutions. Foot Process Width (FPW) MAE denotes the mean per-image pixel error $|\text{fpw}_\text{pred}-\text{fpw}_\text{GT}|$ scaled to $1024\times 1024$. Bold denotes the best result and underline denotes the second-best result among non-ablation methods at the same resolution.}
  \label{tab:ft_results}
  \centering 
  \begin{tabular*}{\textwidth}{@{\extracolsep{\fill}} lcccccc @{}}
    \toprule
    Method & Res. & FT Img/s & Mem. & Slits IoU $(\uparrow)$ & mIoU $(\uparrow)$ & FPW MAE $(\downarrow)$ \\
    \midrule
    \multirow{3}{*}{\shortstack[l]{ViT-MAE}} 
    & 512 & $37$ & $6.7$ GB & $\underline{.447 \pm .008}$ & $\underline{.606 \pm .004}$ & $21.18 \pm 1.22$ \\
    & 768 & $11$ & $14.5$ GB & $\underline{.488 \pm .003}$ & $\mathbf{.627 \pm .001}$ & $19.69 \pm 2.00$ \\
    & 1024 & $4$ & $25.4$ GB & $.494 \pm .009$ & $\underline{.630 \pm .005}$ & $\underline{19.26 \pm 1.12}$\\
    \midrule
    \multirow{3}{*}{\shortstack[l]{SimMIM}}
    & 512 & $82$ & $4.8$ GB & $.414 \pm .002$ & $.575 \pm .007$ &  $\mathbf{19.87 \pm 1.30}$ \\
    & 768 & $37$ & $10.4$ GB & $.478 \pm .005$ & $.616 \pm .002$ & $\mathbf{17.58 \pm 1.43}$ \\
    & 1024 & $21$ & $18.0$ GB & $\underline{.506 \pm .003}$ & $.628 \pm .002$ & $19.70 \pm 0.59$ \\
    \midrule
    \multirow{3}{*}{HiViT} 
    & 512 & $81$ & \textbf{4.1 GB} & $.416 \pm .004$ & $.588 \pm .004$ & $24.93 \pm 1.34$ \\
    & 768 & $36$ & \underline{8.6 GB} & $.471 \pm .003$ & $.616 \pm .004$ & $22.49 \pm 0.23$\\
    & 1024 & $18$ & \underline{14.9 GB} & $.480 \pm .008$ & $.623 \pm .003$ &  $19.54 \pm 0.86$ \\
    \midrule
    \multirow{3}{*}{GreenMIM} 
    & 512 & $42$ & $4.7$ GB & $.415 \pm .005$ & $.562 \pm .005$ & $31.45 \pm 2.00$\\
    & 768 & $22$ & $9.8$ GB & $.450 \pm .005$ & $.581 \pm .004$ & $20.49 \pm 2.43$ \\
    & 1024 & $12$ & $17.1$ GB & $.480 \pm .002$ & $.597 \pm .002$ & $21.69 \pm 2.13$\\
    \midrule
    \multirow{3}{*}{MixMAE} 
    & 512 & $82$ & \underline{4.3 GB} & $.399 \pm .001$ & $.562 \pm .002$ & $28.10 \pm 1.20$ \\
    & 768 & $38$ & $8.7$ GB & $.459 \pm .002$ & $.596 \pm .001$ & $21.24 \pm 1.55$\\
    & 1024 & $21$ & $15.3$ GB & $.473 \pm .004$ & $.607 \pm .003$ & $20.45 \pm 0.65$\\
    \midrule
    nnU-Net V2 & $1024$ & -- & $22$ GB & $.444 \pm .030$ & $.583 \pm .020$ & $31.60 \pm 2.09$ \\
    \midrule
    \multirow{3}{*}{\shortstack[l]{\textbf{AFFMAE}\\\textbf{(Ours)}}} 
    & 512 & $82$ & $6.2$ GB & $\mathbf{.459 \pm .002}$ & $\mathbf{.608 \pm .002}$ & $\underline{21.16 \pm 1.07}$\\
    & 768 & $36$ & \textbf{7.9 GB} & $\mathbf{.490 \pm .005}$ & $\underline{.622 \pm .003}$ & $\underline{18.61 \pm 2.11}$\\
    & 1024 & $20$ & \textbf{13.4 GB} & $\mathbf{.514 \pm .006}$ & $\mathbf{.633 \pm .002}$ & $\mathbf{15.97 \pm 0.92}$\\
    \midrule
    \multirow{3}{*}{\shortstack[l]{\textit{Ablation} \\ ViT-MAE\\Point Dec.}}
    & 512 & $18$ & $15.1$ GB & $.478 \pm .006$ & $.615 \pm .004$ & $19.31 \pm 0.79$ \\
    & 768 & -- & OOM & -- & -- & -- \\
    & 1024 & -- & OOM & -- & -- & -- \\
    \bottomrule
  \end{tabular*}
\end{table*}

\begin{table}[tb]
  \caption{Pre-training throughput (images/sec) FLOPs, total time, and memory at the $512 \times 512$ resolution with a shared batch size of 32.}
  \label{tab:pt_efficiency}
  \centering
  \begin{tabular}{@{}lcccc@{}}
    \toprule
    Method & GFLOPs & Mem. (GB) & Img/s  \\
    \midrule
    ViT-MAE & 274.5 & 29.7 & 76  \\
    MixMAE & 126.3 & 20.1 & 146  \\
    SimMIM & 119.3 & 27.7 & 111 \\
    HiViT & 96.7 & 11.7 & 318 \\
    GreenMIM & 71.3 & 11.9 & 76 \\
    \textbf{AFFMAE (Ours)} & \textbf{58.7} & \textbf{14.5} & \textbf{151} \\
    \bottomrule
  \end{tabular}
\end{table}

\begin{figure}[htbp]
    \centering
    \includegraphics[width=0.75\linewidth]{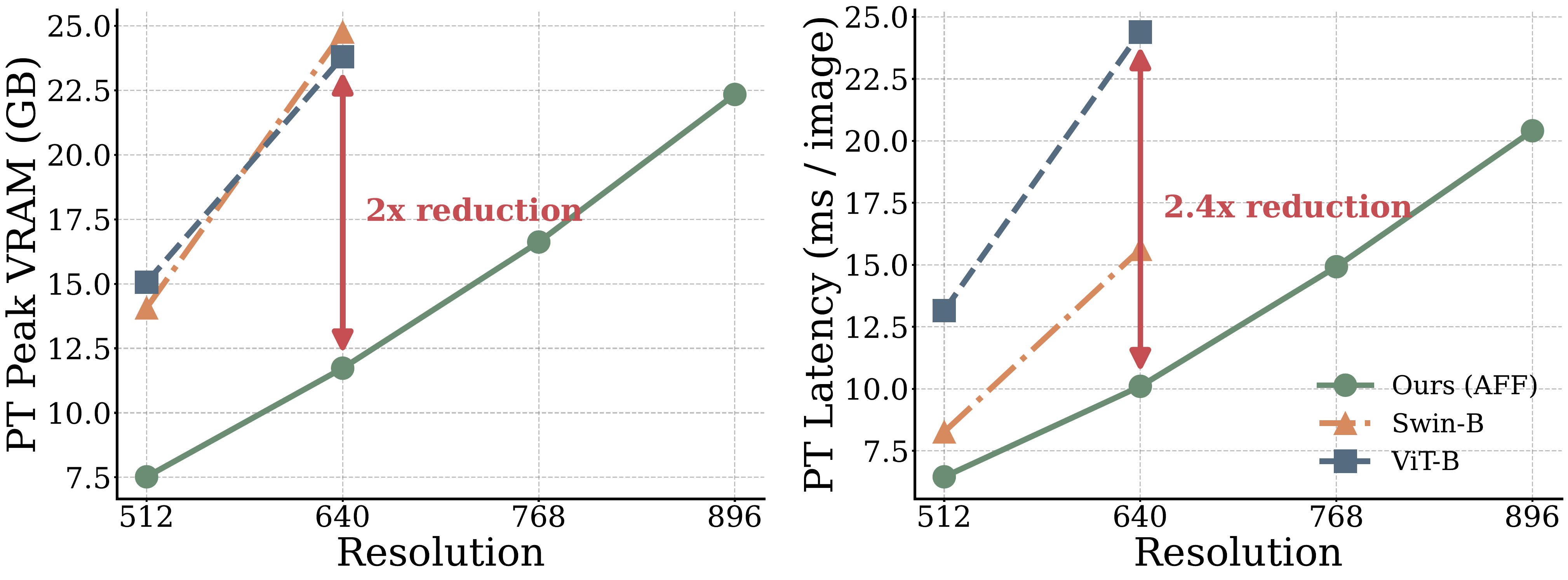}
    \caption{Pre-training computational scaling (batch-size 16) with input resolution. ``PT'' means Pre-Training}
    \label{fig:scaling}
\end{figure}

\textbf{Pre-Training Efficiency and Scalability.} As shown in Table \ref{tab:pt_efficiency}, AFFMAE achieves a pre-training throughput of 151 images/sec, representing a 2x speedup over MAE (76 images/sec) and 1.4x over SimMIM (111 images/sec), at a computational footprint of just  58.7 GFLOPs. More critically, AFFMAE reduces peak pre-training memory to 14.5 GB, a 2x reduction compared to the 29.7 GB and 27.7 GB required by MAE and SimMIM at 32 batch size, respectively. AFFMAE enables efficient model pre-training on a single consumer-grade RTX 5090, whereas dense baselines typically require server-scale infrastructure to be trained at reasonable speeds. Furthermore, natural image models are often pre-trained at lower resolutions and fine-tuned at higher ones, medical imaging relies on high-frequency structures that are destroyed by image downscaling. Pre-training directly at high resolutions is therefore desirable. We reduce the batch size to 16 for this scalability test, as both baselines go Out-of-Memory (OOM) past the 512 resolution at a batch size of 32. Even under this load, MAE and SimMIM fail beyond the $640 \times 640$ resolution on a 32GB GPU, while AFFMAE scales, maintaining latency and consuming 22.3 GB of VRAM at 896 (Figure \ref{fig:scaling}).

\begin{figure}[tb]
    \centering
    \begin{subfigure}[t]{\textwidth}
        \vspace{0pt}
        \centering
        \includegraphics[height=5.0cm, keepaspectratio]{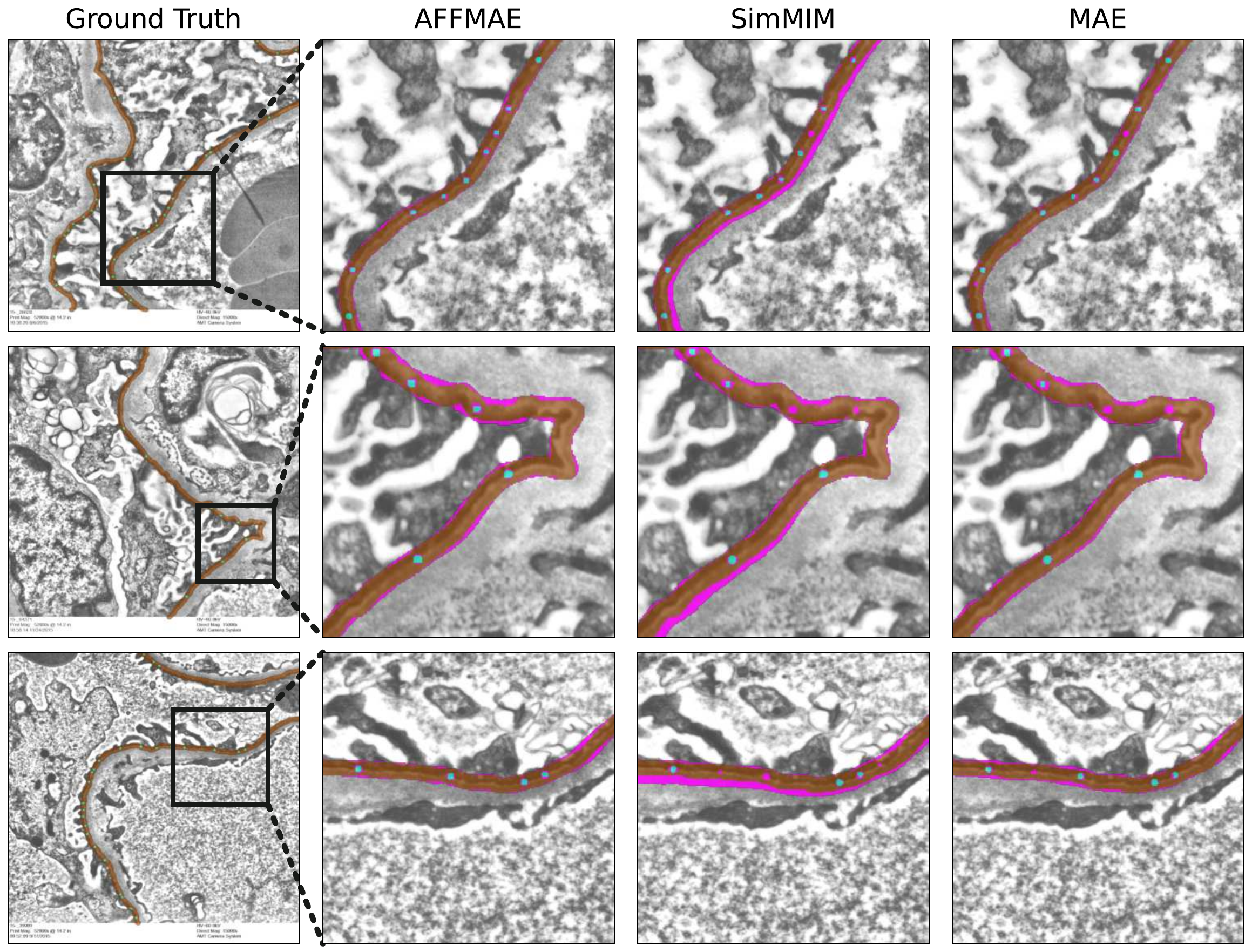}
        \caption{Qualitative segmentation comparison on the FPW dataset with magnified predictions from AFFMAE, SimMIM, and MAE. Dark orange is the PGBMI class, teal is the slits class and segmentation errors are highlighted in pink.}
        \label{fig:qualitative_comparison}
    \end{subfigure}
    \hfill
    \caption{Qualitative evaluation of downstream segmentation.}
    \label{fig:combined_qualitative}
\end{figure}

\textbf{Downstream Finetuning Performance.}
Table \ref{tab:ft_results} details the end-to-end fine-tuning performance across multiple input resolutions. Figure \ref{fig:qualitative_comparison} presents a qualitative comparison between models. AFFMAE achieves segmentation mIoU on par with the dense MAE baseline at all resolutions while drastically reducing computational overhead. At the $1024 \times 1024$ resolution, AFFMAE reaches 0.633 mIoU, closely matching MAE's 0.630 mIoU. In contrast, other efficient MAE baselines such as HiViT and SimMiM plateaus at 0.623 mIoU. This performance parity with MAE is achieved at a fraction of the hardware cost. For upstream pre-training, AFFMAE achieves a throughput of 151 images/s, which is more than 2x faster than MAE (76 images/s), 1.4x faster than SimMIM (111 images/s), while matching or exceeding ViT finetuned mIoU. During downstream fine-tuning at the 1024 resolution, AFFMAE yields a training throughput of 20 images/sec compared to MAE's 4 images/sec (5x speedup) while nearly halving peak memory consumption from 25.4 GB to 13.4 GB. We evaluate AFFMAE against various public datasets in the supplementary materials and observe similar competitive performance at a fraction of the computation. 

\begin{figure}[tb]
    \centering
    \begin{subfigure}[t]{0.8\textwidth}
        \vspace{0pt}
        \centering
        \includegraphics[height=5.0cm, keepaspectratio]{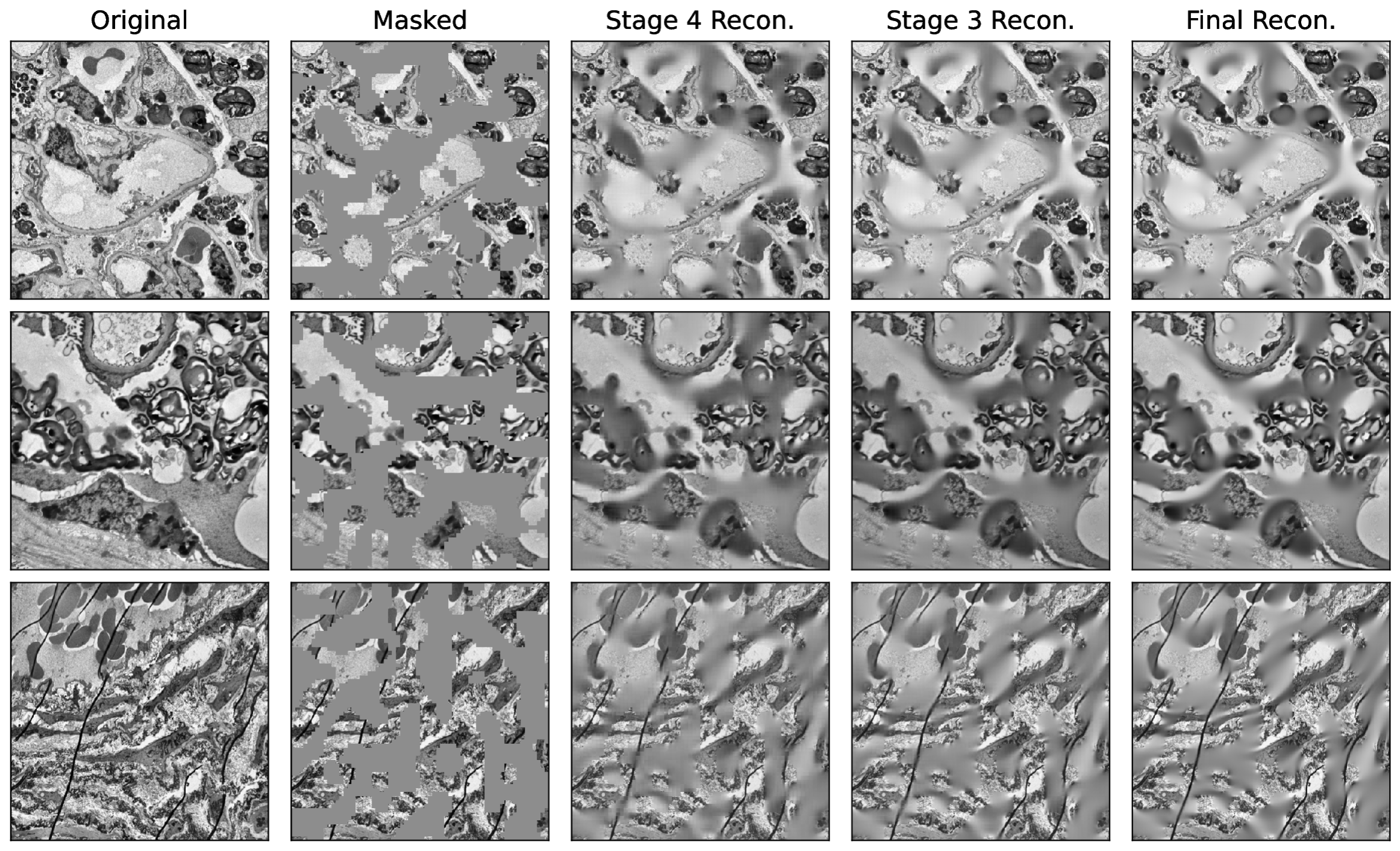}
        \caption{Qualitative reconstruction visualizations of AFFMAE from various stages of the encoder.}
        \label{fig:reconstruction_example}
    \end{subfigure}
    \hfill
    \begin{subfigure}[t]{0.9\textwidth}
        \centering
        \includegraphics[width=\linewidth]{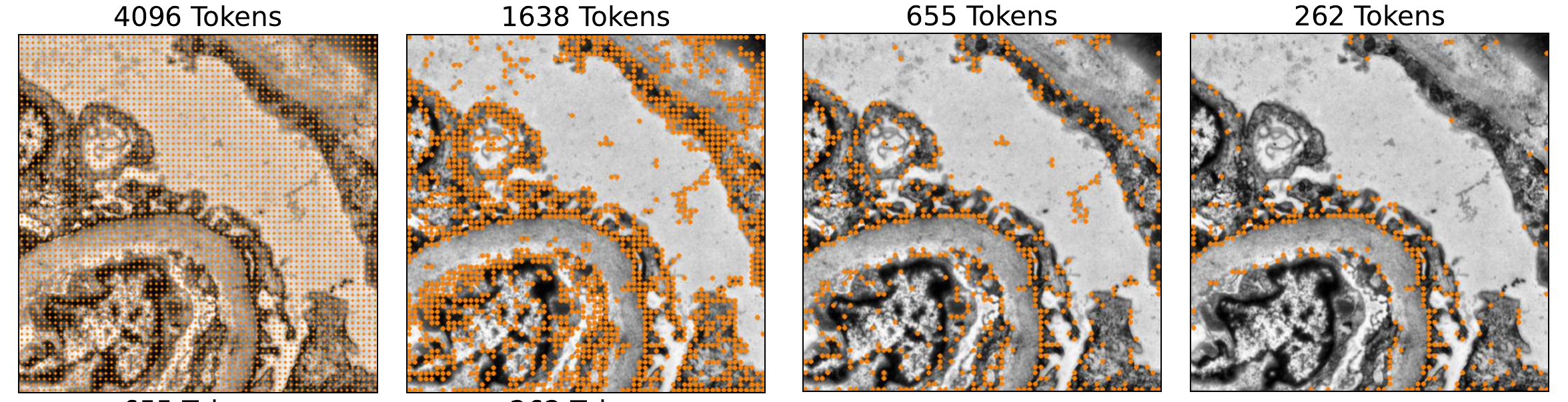}
        \caption{Token merging locations learned by AFFMAE during fine-tuning.}
        \label{fig:token_locations}
    \end{subfigure}
    \caption{Qualitative reconstruction and token location visualizations.}
    \label{fig:combined_reconstruction_tokens}
\end{figure}

\textbf{Segmenting Fine-Grained Structures.} A critical failure point for hierarchical models like Swin is their uniform downsampling on a rigid grid. As shown in Table \ref{tab:ft_results}, hierarchical models like MixMAE, SimMiM, HiViT, and GreenMiM struggle heavily on the small Filtration Slits class, yielding only $.399-.415$ IoU compared to MAE ($.447$ IoU) and AFFMAE ($.459$ IoU) at $512$px. Due to Swin's grid-based patch merging uniformly halving the spatial resolution at each stage, the localized features of these thin slits are irreparably diluted into neighboring patches. Conversely, AFFMAE merges tokens dynamically based on information density rather than a fixed lattice, allowing it to retain high token density around informative regions such as the slits at its sparsest stages (see Figure \ref{fig:token_locations}). This allows AFFMAE to outperform the MAE baseline on the slits class across all resolutions, yielding a Slits IoU of 0.514 at the $1024$ resolution. Additional segmentation results are included in the supplementary materials.

\section{Discussion and Limitations}
\paragraph{Explainability.}
Beyond enabling efficient high-resolution training, AFF offers another form of interpretability through  image-dependent token locations. In conventional hierarchical backbones, downsampling is grid-aligned and deterministic, so token coordinates are fixed and carry little explicit information about the image. In AFF, token merging is adaptive: tokens are retained and concentrated in regions of high information content, and their locations change with the input. As a result, the learned token trajectories can be visually inspected, and potentially used as downstream task-relevant signals. For FPW segmentation, Figure~\ref{fig:token_locations} shows that by the deepest stage, tokens concentrate along the GBM boundary and largely disappear from the visually similar parietal membrane side, providing qualitative evidence that the model allocates token capacity to structures that drive slit detection. This suggests a promising direction for downstream measurement tasks that directly leverage token locations, such as GBM width or slit spacing, alongside dense predictions in an end-to-end framework.

\paragraph{Extension to volumetric imaging.}
Our work focuses on 2D TEM because it reflects the composition of our EM archive and the current clinical FPW measurement workflow, where 2D sections are far more abundant and routinely used for measurement. However, AFFMAE is not architecturally restricted to 2D. Its core operations---explicit token coordinates, balanced cluster attention, adaptive token merging, and point-based virtual sampling---extend naturally from image-plane tokens $\mathbf{x}_i\in\mathbb{R}^2$ to volumetric tokens $\mathbf{x}_i\in\mathbb{R}^3$. In a 3D setting, cluster-attention neighborhoods become local spatial clusters in voxel or physical coordinate space, anchor selection and token merging operate over volumetric neighborhoods, and decoder offsets/reference points become 3D queries that gather nearby visible tokens from multi-scale volumetric feature sets. This extension is appealing for volumetric microscopy since token counts grow cubically with input side length $n$: for a cubic volume, dense global attention scales as $O(n^6)$ rather than the $O(n^4)$ scaling of 2D inputs. A 3D AFF-style encoder could therefore preserve token density near thin membranes, slits, organelle boundaries, or other sparse high-frequency structures while merging less informative regions. We view volumetric AFFMAE as a natural future direction, with open challenges in efficient balanced clustering, 3D neighborhood lookup, anisotropy-aware masking, and optimized kernels for volumetric token sets.

\paragraph{Limitations.}
A limitation of AFF is that its irregular token set complicates low-level efficiency compared to dense grid indexing and can increase memory traffic. Although AFF substantially reduces FLOPs, these reductions do not always translate into proportional wall-clock speedups. Grid-based models benefit from simple addressing and neighborhood lookups, whereas AFF relies on neighborhood queries, KNN operations, and gather/scatter patterns that can become bandwidth-limited. We mitigate this by caching neighborhood tables for encoder stages and for the decoder's virtual token sampling, where neighbor queries are invoked repeatedly. However, at very high resolutions, dense lookup tables can grow large enough to exceed cache capacity, making simple grid indexing competitive or faster in some regimes. Improving neighborhood lookup, space-filling strategies, caching, and decoder design will therefore be important for scaling beyond $\sim$1200px inputs and for extending AFFMAE to volumetric data. Further kernel optimization also remains an opportunity to close the gap between theoretical FLOP savings and realized throughput. 

\section{Conclusion}
We introduced AFFMAE, a masking-friendly hierarchical pretraining framework for high-resolution microscopy segmentation on desktop-class hardware. AFFMAE combines adaptive off-grid token merging, a mask-agnostic encoder, a point-based decoder, and efficient mixed-precision Triton kernels to retain MAE-style visible-token pretraining while scaling to high-resolution inputs. Across microscopy segmentation benchmarks, AFFMAE matches ViT-MAE performance, and outperforming Swin based methods with small segmentations, at comparable parameter counts while reducing pre-training time by up to $2\times$, high-resolution fine-tuning time by up to $5\times$, and peak memory by up to $2\times$ compared to standard MAE. These gains make in-domain self-supervised pretraining practical for laboratories with large private archives but limited access to server-scale infrastructure. More broadly, AFFMAE suggests that learned adaptive token allocation can preserve fine biological structures without paying the full cost of dense global computation, providing a promising path toward scalable pretraining for high-resolution 2D, volumetric, and time-resolved biomedical imaging.

\section*{Acknowledgements}
We thank Dr. Michael Mauer for generously providing access to an extensive electron microscopy archive used for pretraining in this study. We are grateful for his support in making this research possible.

%
%
\bibliographystyle{splncs04}
\bibliography{main}


\title{Supplemental Materials for AFFMAE} 

\titlerunning{AFFMAE}

\section{Implementation Details, Additional Benchmarks, and Additional Qualitative Results}
\begin{figure}[htbp]
    \centering
    \includegraphics[width=1.0\linewidth]{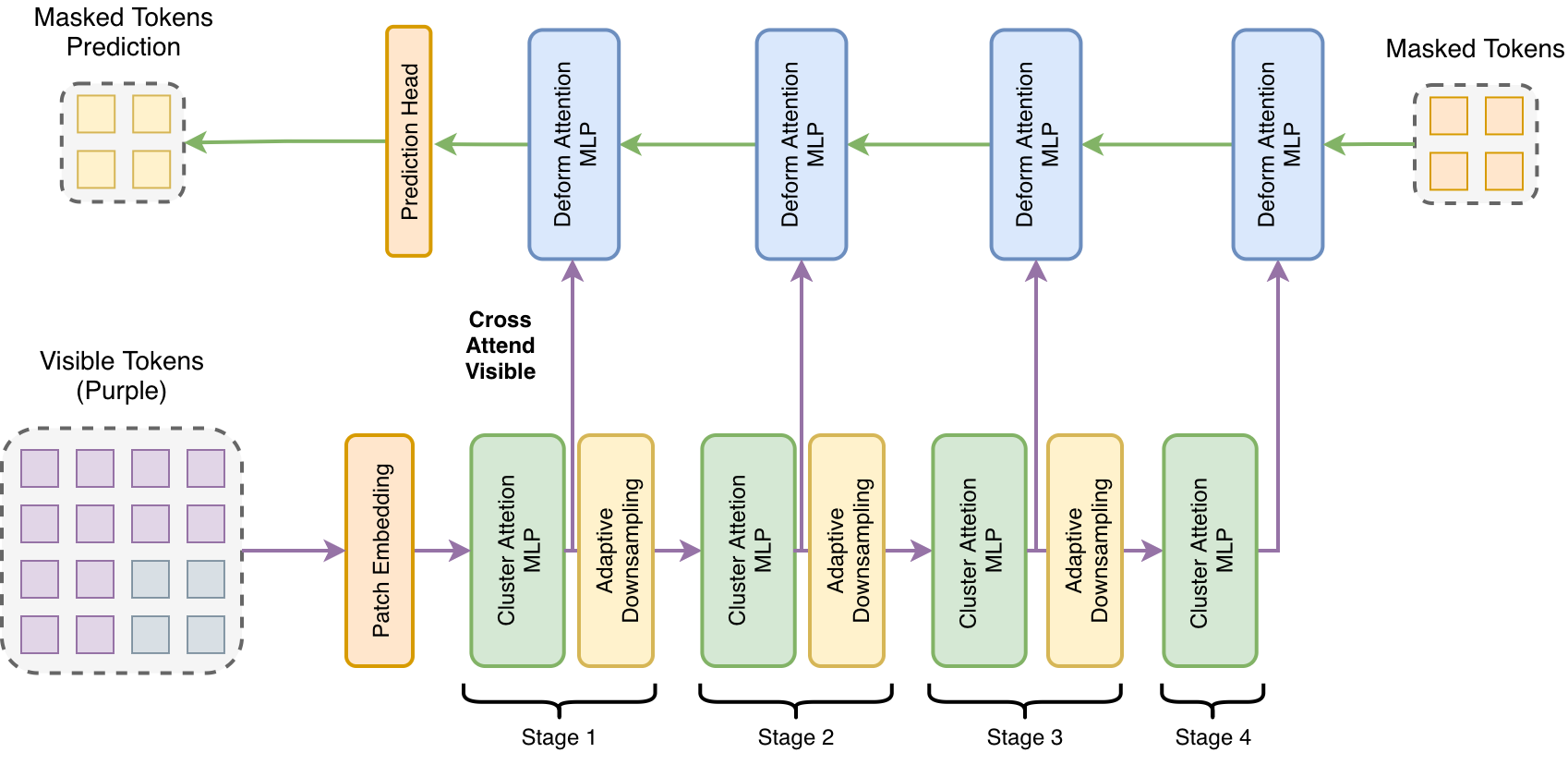}
    \caption{\textbf{Diagram of the AFFMAE framework.} The AFF encoder consists of four stages, each with $N$ repeating cluster transformer blocks. Intermediate features are collected before adaptive downsampling/token merging, except in the final stage where there is no token merging step. In the decoder, mask tokens with absolute positional embeddings serve as queries to cross-attend to encoder intermediate features via deformable cross-attention. Note that the auxiliary reconstruction heads utilized by Deep Supervision are not pictured here.}
    \label{fig:supp_framework}
\end{figure}

\begin{figure}
    \centering
    \includegraphics[width=0.9\linewidth]{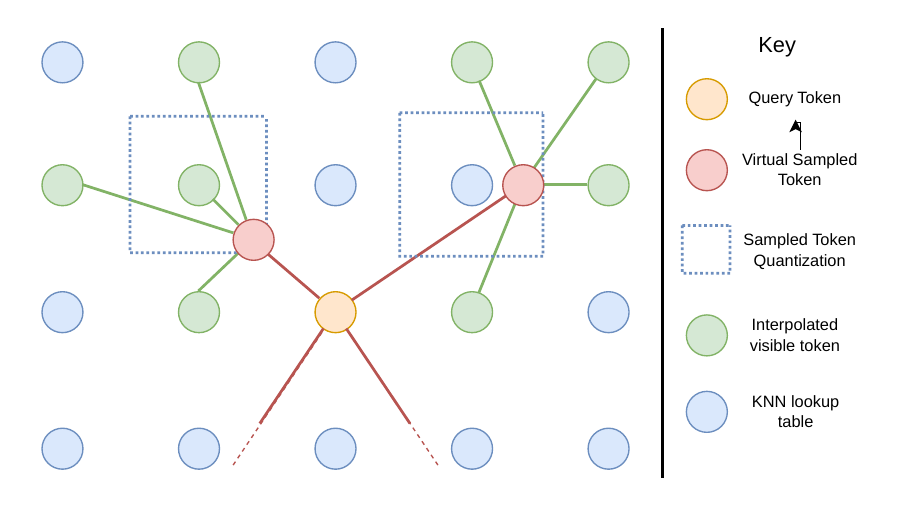}
    \caption{Implementation of AFFMAE's point-based KNN decoder attention using a quantized cache KNN grid.}
    \label{fig:knn_grid}
\end{figure}

\subsection{Point-based decoder}
To speed up the decoder, we aimed to remove one major pain point of the point-based decoder: the KNN lookup. When performing attention on 4096 points, as in stage 1, each query point attends to 4 virtual sampled tokens, and each sampled token requires a KNN lookup. This requires a dense lookup for each sampled point separately. This results in \(Q \times 4\) KNN lookups per point-decoder block. To simplify the kernel, we propose using a cached KNN table for masked tokens. Specifically, we compute a quantized version of the table in which each element contains the indices of the 4 nearest visible-token neighbors, and cache it. This is especially useful for masked-token self-attention, since the points do not change position in deeper blocks. As illustrated in Figure~\ref{fig:knn_grid}, we gather neighbors for each sample point by rounding its location to the nearest KNN grid point. This table can be reused throughout the decoder and makes writing a fused-attention decoder much simpler. We observed a \(1.5\times\) speedup in forward time and a \(1.4\times\) speedup in backward time when using the KNN lookup table approach (Table~\ref{tab:speedup}).

\begin{table}[tb]
  \caption{Pixel decoder speed improvement with new kernels. Times are estimated with a batch size of 48, 2048 mask tokens, and 30 trials.}
  \label{tab:speedup}
  \centering
  \begin{tabular}{lccc}
    \hline
    \textbf{Mode} & \textbf{Forward Time (ms)} & \textbf{Backward Time (ms)} & \textbf{Speedup} \\
    \hline
    Standard      & $94.00 \pm 0.54$ & $227.21 \pm 1.38$  & $1\times$\\
    KNN Cache & $62.75 \pm 2.47$ & $160.19 \pm 2.73$ & $1.418\times$\\
    \hline
  \end{tabular}
\end{table}

\subsection{Architecture Configurations}
\textbf{ViT architecture.} The ViT model uses a patch size of 8. The encoder features an embedding dimension of 512, a depth of 18, and 8 attention heads. The decoder consists of an embedding dimension of 384, a depth of 4, and 6 attention heads. The MLP ratio is set to 4.0, and LayerNorm is used as the normalization layer throughout the model.

\textbf{Swin architecture.} We select a patch size of 4 to adhere to the standard Swin hierarchical design, which establishes the initial feature maps at a quarter of the input resolution. The network uses an embedding dimension of 128, and a window size of 16. The block depths across the four stages are $[2, 2, 10, 2]$, and the corresponding number of attention heads are $[4, 8, 8, 16]$. The MLP ratio is set to 4.0. We use the default masking strategy with a mask patch size of 32. LayerNorm is used as the normalization layer.

\textbf{AFF architecture.} The AFF model operates on a patch size of 8. The embedding dimensions for the four stages are $[128, 256, 512, 768]$. A linear projection is applied at the end of each token merging block to increase the embedding dimension. The depths for the respective stages are $[3, 4, 16, 2]$. We set the neighborhood size to 64 across all stages, the cluster size to 8, and the MLP ratio to 3. The decoder features an embedding dimension of 384, a depth of 4, and 6 attention heads. LayerNorm is used as the normalization layer.

\textbf{nnU-Net V2 architecture.} We use the nnU-Net ResEnc L configuration, which is the default recommended architecture for a GPU with 32GB of VRAM. The framework is executed following the official instructions without modification. Note that nnU-Net employs sliding-window inference with overlapping patches, automated post-processing, and ensembling, which are techniques not utilized by our other baseline models.

\subsection{Pre-Training Configurations}
\begin{table}
  \caption{Pre-training FLOPs, total time, and memory at the $512 \times 512$ resolution for models pretrained on 5090.} 
  \label{tab:supp_pt_efficiency}
  \centering
  \begin{tabular}{@{}lcccc@{}}
    \toprule
    Method & GFLOPs & Mem. (GB) & PT Hours \\
    \midrule
    ViT-MAE & 274.5 & 29.7 & 281 \\
    SimMIM & 119.3 & 27.7 & 191 \\
    \textbf{AFFMAE (Ours)} & \textbf{58.7} & \textbf{14.5} & \textbf{138} \\
    \bottomrule
  \end{tabular}
\end{table}
We pre-train the models using AdamW with $\beta_1 = 0.883$ and $\beta_2 = 0.935$, as suggested by MAE \cite{he2022masked}. The learning rate follows a cosine annealing schedule with a base learning rate of 3.5e-4, a minimum learning rate of 1e-6, and a linear warmup period of 10,000 steps. Weight decay is set to 0.05. For image preprocessing, we apply Contrast Limited Adaptive Histogram Equalization (CLAHE) with a clip limit of 4 and a tile grid size of 8x8. Images are subsequently normalized using the dataset's mean and standard deviation. No additional data augmentations are applied during the pre-training phase. Pretraining times on an RTX 5090 machine (ViT-MAE, AFFMAE, and SimMiM) are presented in Table~\ref{tab:supp_pt_efficiency}. Note, the other compared baselines (GreenMiM, HiViT, and MixMAE) were ran on a H100 and do not have easily comparable pretraining time to 5090.

\subsection{Segmentation on the FPW Dataset}
For the FPW segmentation experiments, all MAE, SimMIM, and AFFMAE models were trained for 400 epochs with a base learning rate of $1 \times 10^{-4}$, a minimum learning rate of $10^{-6}$, and 25 warmup epochs. WE utilized a combined Binary Cross-Entropy and Dice loss with a class weighting of $[0.2, 2.0, 3.0]$ for the background class, PGBMI class, and the slits class, respectively. Data augmentation included affine transformations, photometric distortion, and elastic distortion. We applied a layer-wise learning rate decay of 0.6. As the pre-training reconstruction head is single-channel, we initialized a new segmentation head with an output dimension corresponding to the number of target classes.

\subsection{Segmentation on the Lucchi$++$ Dataset}

For the Lucchi$++$ segmentation experiments, all MAE, SimMIM, and AFFMAE models were trained for 200 epochs using the same augmentation pipeline as in the other fine-tuning experiments. We used AdamW with a base learning rate of \(1.5\times10^{-4}\), weight decay \(7.5\times10^{-3}\), and a layer-wise learning rate decay of 0.75. We only use Dice as the loss function. As shown in Table~\ref{tab:public_benchmarks_comparison}, MAE achieved the best performance on Lucchi$++$ with an mIoU of \textbf{0.8840}, narrowly outperforming AFFMAE at \(0.8829\), while SimMIM trailed behind at \(0.8266\).

\subsection{Segmentation on the 2012 ISBI EM Challenge Dataset}
For the ISBI EM Challenge segmentation experiments, all MAE, SimMIM, and AFFMAE models were trained and evaluated at an input resolution of \(512 \times 512\) for 200 epochs. Across all runs, we used a consistent optimization setup with learning rate \(2\times10^{-4}\), weight decay \(10^{-3}\), 50 warmup epochs, and a layer decay rate of 0.7. For the ViT-based baselines, the best configuration used a combined Dice + BCE segmentation loss, while for AFFMAE the best configuration additionally enabled deep supervision. As shown in Table~\ref{tab:public_benchmarks_comparison}, AFFMAE achieved the best ISBI performance with an mIoU of 0.8661, slightly outperforming MAE (\(0.8649\)) and SimMIM (\(0.8488\)). These results indicate that AFFMAE works competitively on other larger segmentation masks, not just foot process width estimation.

\subsection{Segmentation on the Kasthuri++ Dataset}
\begin{figure}
    \centering
    \includegraphics[width=0.9\linewidth]{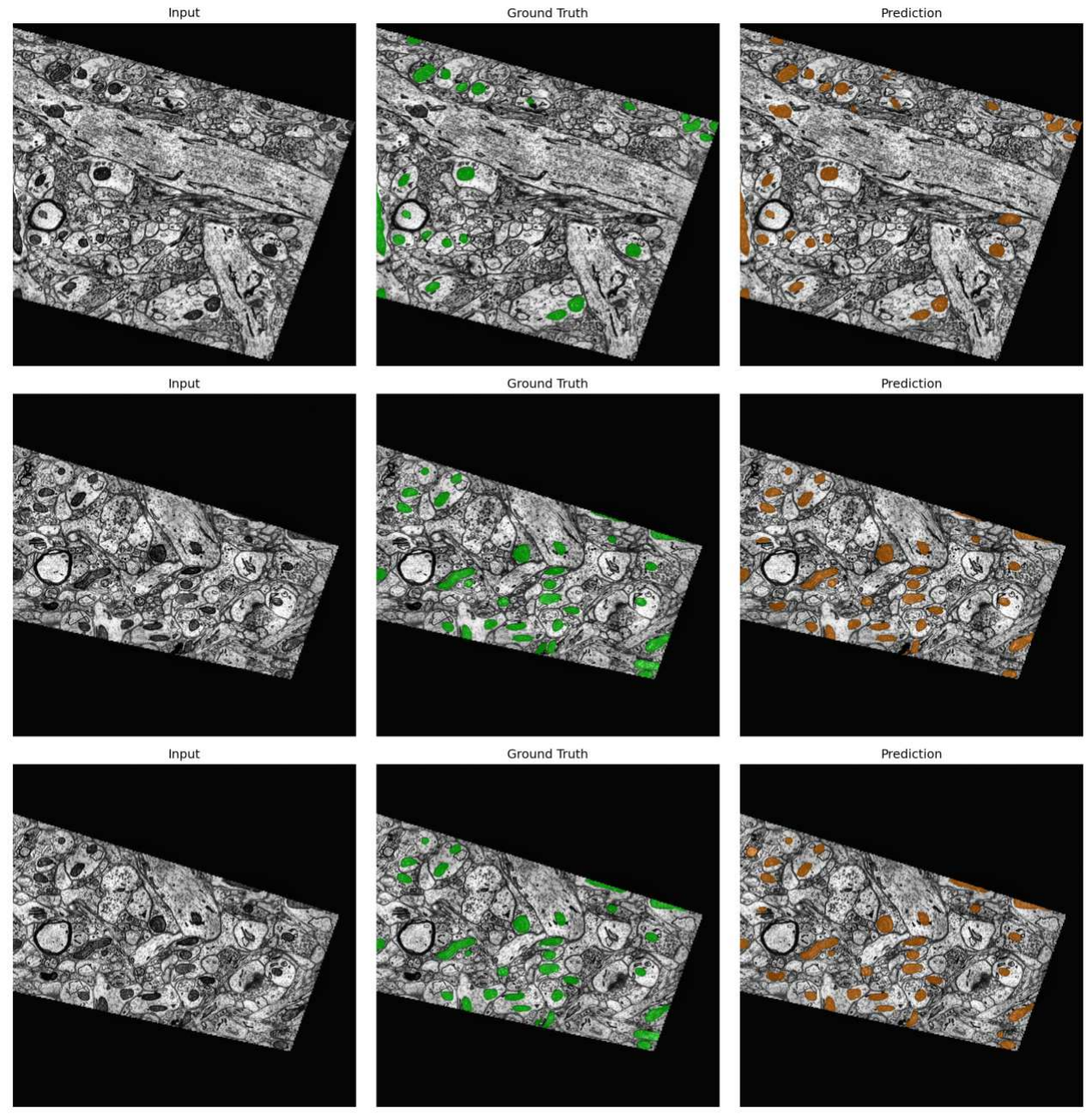}
    \caption{Random sample of Kasthuri++ AFF predictions compared to ground truth.}
    \label{fig:kasthuri_pred}
\end{figure}

\begin{figure}
    \centering
    \includegraphics[width=0.9\linewidth]{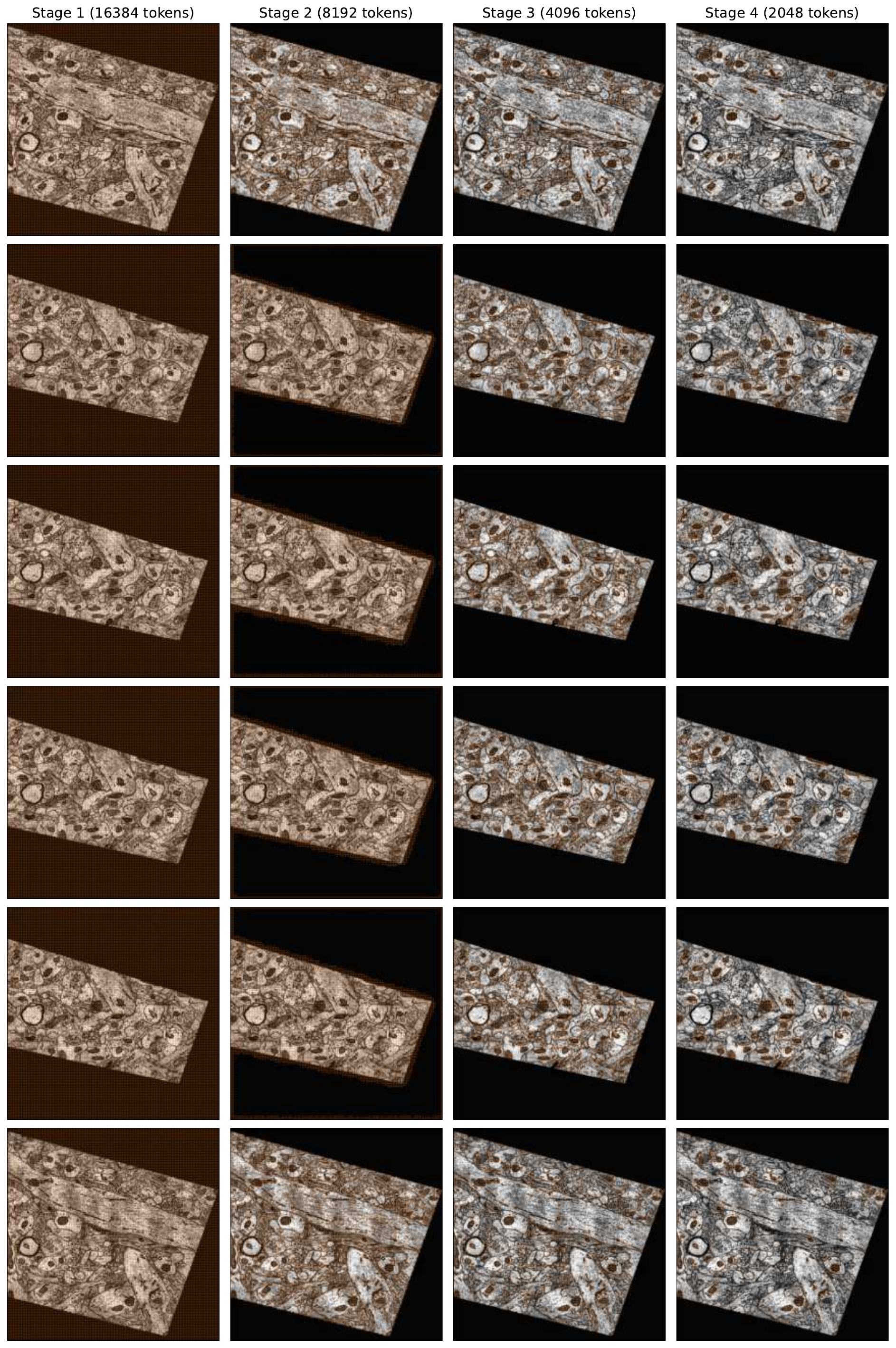}
    \caption{Demonstration of adaptive downsampling on Kasthuri++ dataset.}
    \label{fig:kasthuri_tokens}
\end{figure}

For the Kasthuri++ segmentation experiments, all MAE, SimMIM, and AFFMAE models were trained and evaluated at an input resolution of \(1024 \times 1024\). We used AdamW with a learning rate of \(2\times10^{-4}\), weight decay of \(10^{-3}\), 75 warmup epochs, and a layer-wise learning rate decay of 0.7. For these experiments, we used a Dice-only segmentation objective for all methods, while AFFMAE additionally enabled deep supervision. As shown in Table~\ref{tab:public_benchmarks_comparison}, AFFMAE achieved the strongest performance on Kasthuri++ with an mIoU of \textbf{0.8870}, outperforming both MAE (\(0.8818\)) and SimMIM (\(0.8507\)). For qualitative results, Figure~\ref{fig:kasthuri_pred} shows a random batch of AFFMAE segmentations compared with ground truth. Interestingly, the adaptive downsampling behavior shown in Figure~\ref{fig:kasthuri_tokens} indicates that after the first stage, most tokens corresponding to areas outside the tissue section are removed, while tokens on the specimen itself are retained. This suggests that the adaptive token selection mechanism is able to focus computation on the biologically relevant image content while discarding large uninformative background regions.

\begin{table}[tb]
\centering
\caption{Segmentation performance (mIoU) across public electron microscopy benchmarks.}
\label{tab:public_benchmarks_comparison}
\setlength{\tabcolsep}{12pt} 
\begin{tabular}{lccc}
  \toprule
  Method & Lucchi++ & Kasthuri++ & ISBI EM Challenge \\
  \midrule
  MAE & \textbf{0.8840} & 0.8818 & 0.8649 \\
  SimMIM & 0.8266 & 0.8507 & 0.8488 \\
  AFFMAE (Ours) & 0.8829 & \textbf{0.8870} & \textbf{0.8661} \\
  \bottomrule
\end{tabular}
\end{table}

\begin{figure}[htbp]
    \centering
    \includegraphics[width=0.87\linewidth]{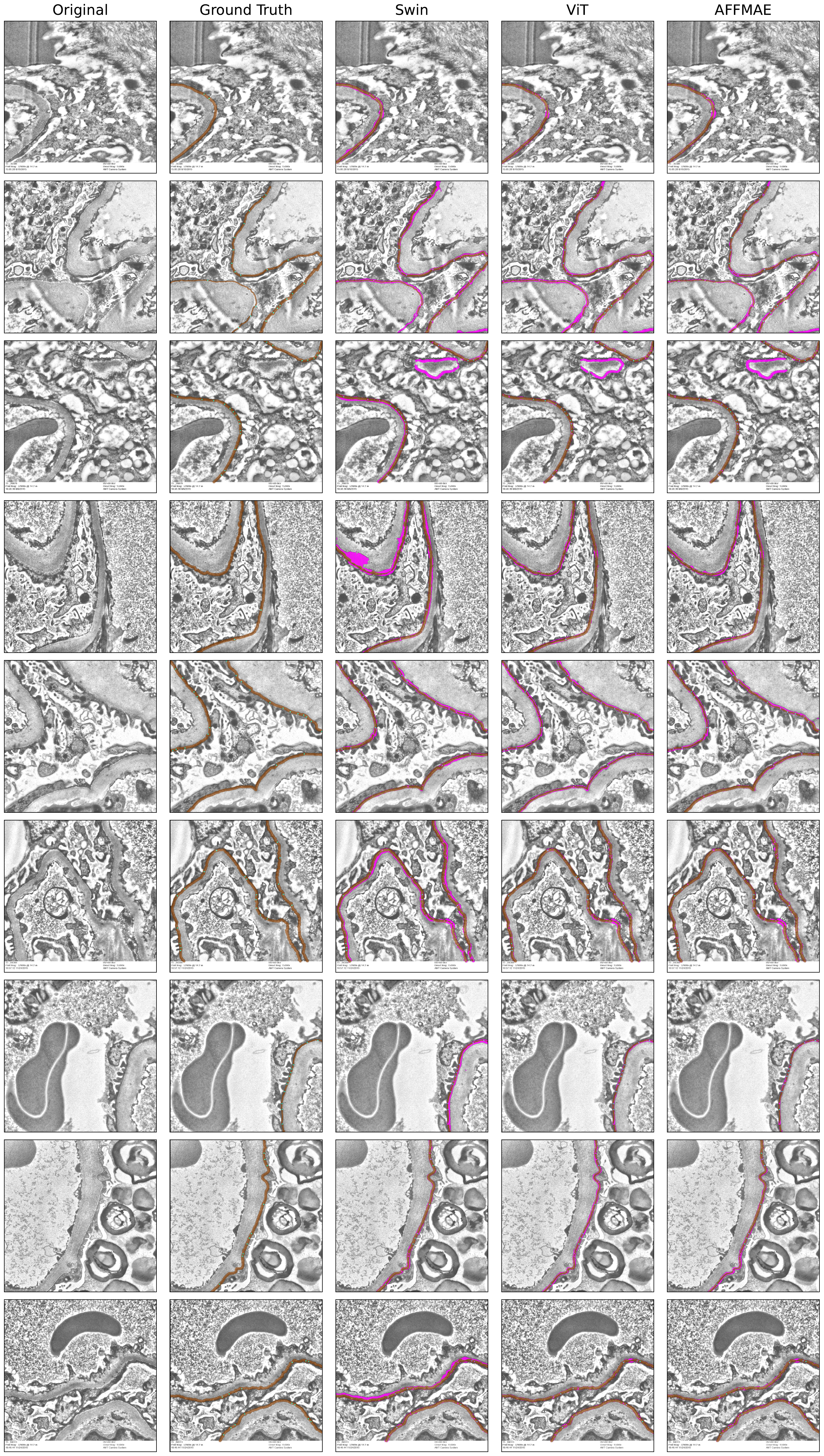}
    \caption{\textbf{Additional Qualitative Segmentation Results.} Brown represents the PGBMI class, teal represents the slits class, while pink represents incorrect segmentation.}
\end{figure}

\begin{figure}[htbp]
    \centering
    \includegraphics[width=0.87\linewidth]{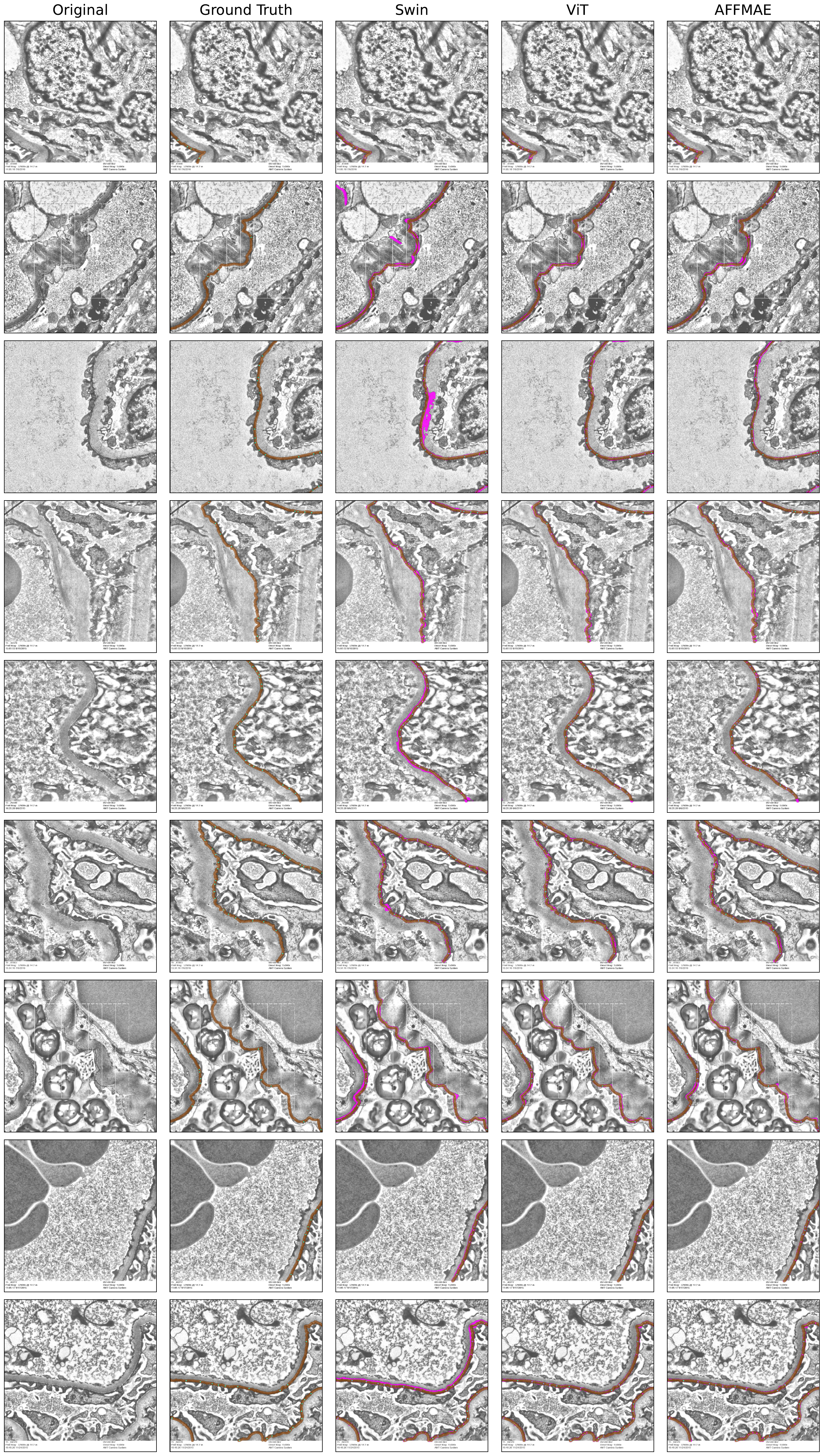}
    \caption{\textbf{Additional Qualitative Segmentation Results.} Brown represents the PGBMI class, teal represents the slits class, while pink represents incorrect segmentation.}
\end{figure}

\begin{figure}[htbp]
    \centering
    \includegraphics[width=0.87\linewidth]{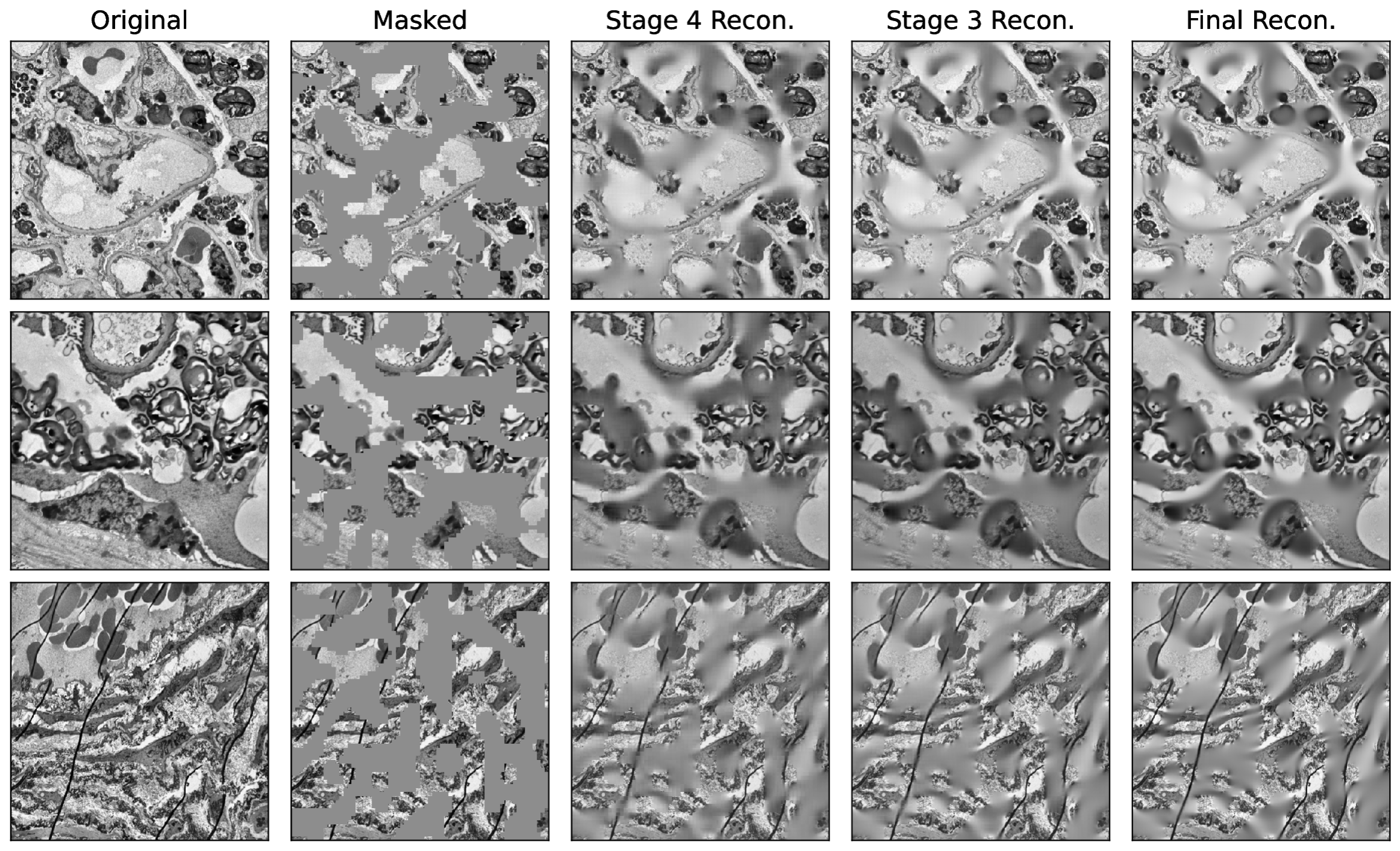}
    \caption{\textbf{Reconstruction Visualizations.} Reconstruction examples from AFFMAE with Perlin masking. Note that "Stage 4 Recon." and "Stage 3 Recon." indicates the reconstruction from the auxiliary heads as part of Deep Supervision.}
\end{figure}

\begin{figure}
    \centering
    \includegraphics[width=0.9\linewidth]{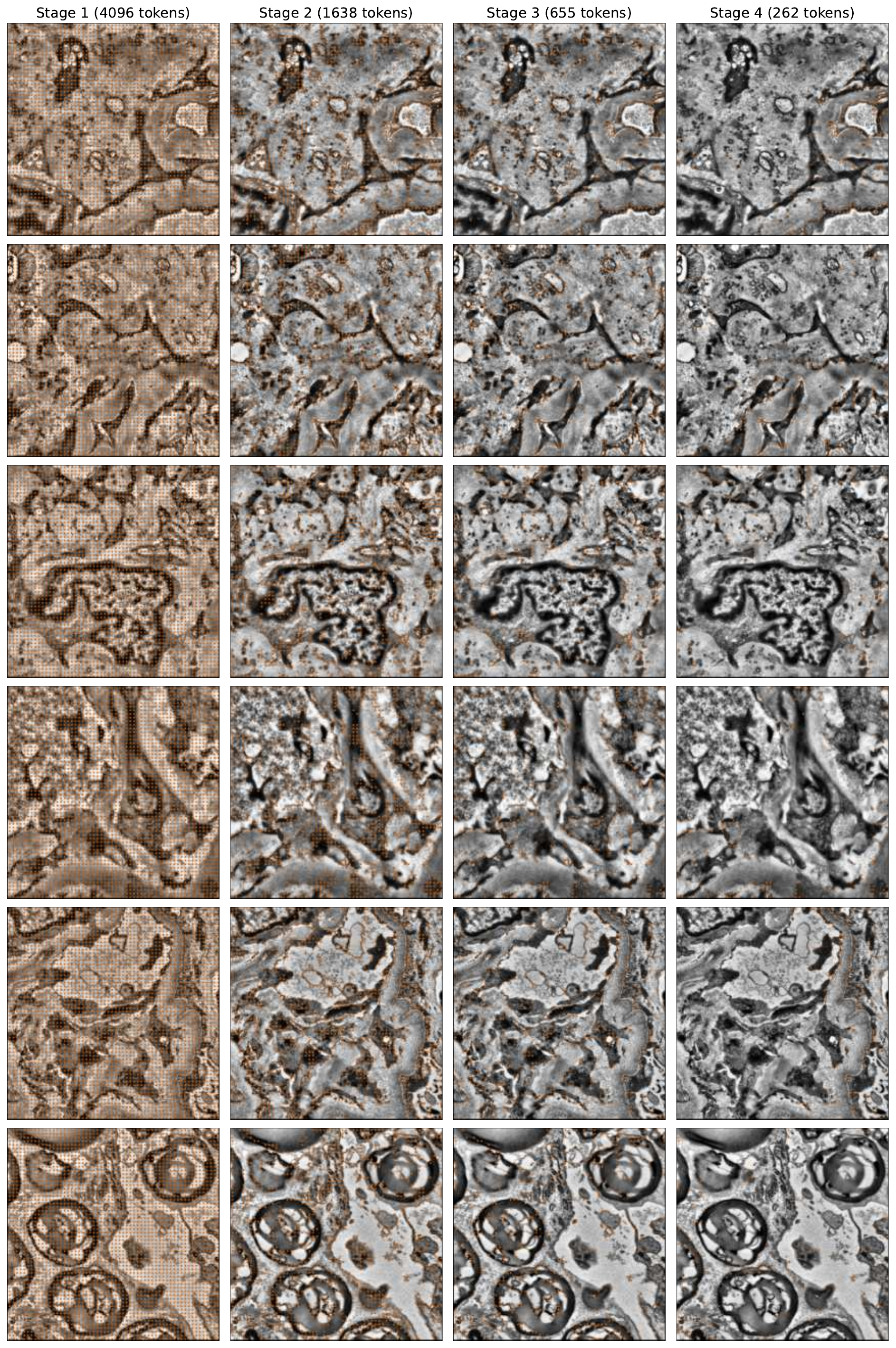}
    \caption{\textbf{Qualitative token placement results.} Each orange token shows the resulting token processed within that stage, before downsampling at a 0.4 downsampling rate.}
    \label{fig:fpw_tokens}
\end{figure}

\end{document}